\documentclass{article}

 \usepackage[eandd,nonanonymous,final]{neurips_2026}

\usepackage[utf8]{inputenc} 
\usepackage[T1]{fontenc}    
\usepackage{hyperref}       
\hypersetup{hidelinks}      
\usepackage{url}            
\usepackage{booktabs}       
\usepackage{amsfonts}       
\usepackage{nicefrac}       
\usepackage{microtype}      
\usepackage{xcolor}         

\usepackage[capitalize,noabbrev]{cleveref}

\usepackage{subcaption}
\usepackage{multirow}
\usepackage{tabularx}
\usepackage{booktabs}
\usepackage{CJKutf8}
\usepackage{ragged2e}
\usepackage{CJKutf8}
\usepackage{multicol}
\usepackage{graphicx}
\usepackage{wrapfig}
\usepackage{ulem}

\title{CS-Dialogue: A 104-Hour Dataset of Spontaneous Mandarin-English Code-Switching Dialogues for Speech Recognition}

\author{
 \textbf{Jiaming Zhou\textsuperscript{1}}, \textbf{Yujie Guo\textsuperscript{1}},  \textbf{Shiwan Zhao\textsuperscript{1}}, \textbf{Haoqin Sun\textsuperscript{1}}, \textbf{Hui Wang\textsuperscript{1}}, \textbf{Jiabei He\textsuperscript{1}},
 \\
   \textbf{Aobo Kong\textsuperscript{1}}, \textbf{Shiyao Wang\textsuperscript{1}}, \textbf{Yong Qin\textsuperscript{1}}\thanks{Yong Qin is the corresponding author.} 
\\
\\
     \textsuperscript{1} College of Computer Science, Nankai University, \\
   \textbf{Correspondence:} \href{mailto:zhoujiaming@mail.nankai.edu.cn}{zhoujiaming@mail.nankai.edu.cn}, \href{mailto:qinyong@nankai.edu.cn}{qinyong@nankai.edu.cn}
}

\begin{document}

\maketitle

\begin{abstract}
Code-switching (CS), the alternation between two or more languages within a single conversation, presents significant challenges for automatic speech recognition (ASR) systems. Existing Mandarin-English code-switching datasets often suffer from limitations in size, spontaneity, and the lack of full-length dialogue recordings with transcriptions, hindering the development of robust ASR models for real-world conversational scenarios.  This paper introduces CS-Dialogue, a novel large-scale Mandarin-English code-switching speech dataset comprising 104 hours of spontaneous conversations from 200 speakers.  Unlike previous datasets, CS-Dialogue provides full-length dialogue recordings with complete transcriptions, capturing naturalistic code-switching patterns in continuous speech.  We describe the data collection and annotation processes, present detailed statistics of the dataset, and establish benchmark ASR performance using state-of-the-art models. Our experiments, using Transformer, Conformer, and Branchformer, demonstrate the challenges of code-switching ASR, and show that existing pre-trained models such as Whisper still have the space to improve. The CS-Dialogue dataset is publicly available for academic research at \url{https://huggingface.co/datasets/BAAI/CS-Dialogue}.
\end{abstract}

\section{Introduction}
Code-switching (CS) refers to the practice of alternating between two or more languages within a single conversation or utterance~\citep{moyer2002bilingual}. It is a common linguistic phenomenon in multilingual communities and occurs in various communication settings, including spoken dialogues, social media, and written texts. The increasing prevalence of code-switching presents significant challenges for automatic speech recognition (ASR) systems, as they must effectively handle complex acoustic and linguistic variations across different languages~\citep{yilmaz2018acoustic}.

Traditional ASR systems, predominantly trained on monolingual data, struggle with code-switched speech due to mismatches in phonetic inventories, syntactic structures, and language switching patterns~\citep{mustafa2022code,zhou2024improving}. These challenges necessitate the development of specialized ASR models and high-quality datasets tailored for code-switching scenarios. Despite recent advancements, existing Mandarin-English code-switching speech corpora remain limited in size, spontaneity, and accessibility, restricting further research and model development.

Table~\ref{tab:datasets} summarizes publicly available Mandarin--English CS datasets. Many corpora~\citep{shen2011cecos,wang2016oc16,TALCS} focus on read or semi-scripted speech, are restricted to narrow domains, lack full dialogue-level transcriptions, or are not freely accessible. Crucially, prior Mandarin--English CS corpora primarily consist of isolated utterances or short segments rather than full dialogues~\citep{10094707}, making it difficult to study conversational dynamics, discourse-level context, and how code-switching unfolds over longer interactions.

To address these gaps, we introduce \textbf{CS-Dialogue}, a large-scale Mandarin--English code-switching speech dataset comprising 104 hours of spontaneous conversations from 200 speakers. Unlike prior work, CS-Dialogue provides full-length dialogue recordings with complete transcriptions, organized by communicative topics and collected under a light-weight temporal guideline that encourages, but does not script, code-switching. This design preserves natural conversational structure while yielding a substantial amount of mixed-language speech, enabling more comprehensive investigations of CS ASR beyond isolated sentences. To the best of our knowledge, CS-Dialogue is the largest publicly available corpus of spontaneous Mandarin--English code-switching dialogues with full transcriptions.

In this paper, we describe the data collection protocol, annotation pipeline, and quality-control procedures, present a detailed characterization of the corpus, and evaluate its impact on ASR performance through a series of baseline and transfer experiments. Our main contributions are:

\begin{itemize}
    \item We construct a large-scale, spontaneous Mandarin-English code-switching speech corpus with full-length dialogue transcriptions, filling the gap of publicly available datasets in this domain.
    \item We detail the data collection and annotation processes, ensuring high transcription accuracy and providing a well-documented resource for future research.
    \item We establish benchmark ASR performance on our dataset using state-of-the-art models, offering insights into the challenges of code-switching ASR.
    \item We publicly release CS-Dialogue under a CC BY-NC-SA 4.0 license for non-commercial academic research, enabling reproducible studies on Mandarin--English code-switching.
\end{itemize}

\begin{table*}[!t]
\centering
\caption{Comparison of Mandarin-English code-switching speech datasets. "Tr." indicates whether transcripts are available, "Avail." specifies whether the dataset is publicly accessible, and "Full-dialogue" denotes whether full-length dialogue recordings and transcriptions are provided.}
\label{tab:datasets}
\resizebox{1.0\linewidth}{!}{
\begin{tabular}{lccccc|c}
\toprule
Dataset & Duration (h) & \#Speakers & Audio Type & Tr. & Avail. & Full-dialogue  \\
\midrule
CECOS~\citep{shen2011cecos}    & 12.1  & 77    & Read           & $\times$     & $\times$   & $\times$ \\
OC16-CE80~\citep{wang2016oc16}   & 80    & 1400+ & Read           & $\checkmark$ & $\times$   & $\times$ \\
ASRU~\citep{shi2020asru}       & 240   & N/A      & N/A              & $\checkmark$ & $\times$   & $\times$ \\
TALCS~\citep{TALCS}          & 587   & 100+  & Online Teaching & $\checkmark$ & $\checkmark$ & $\times$ \\
DOTA-ME-CS~\citep{li2025dota} & 18.54 & 34    & Read    & $\checkmark$ & $\checkmark$ & $\times$ \\
\midrule
SEAME~\citep{SEAME}         & 30    & 157   & Conversation    & $\checkmark$ & Paid  & $\times$ \\
Li et al.~\citep{li2012mandarin} & 36    & N/A      & Conversation    & Partial & $\times$   & $\times$ \\
ASCEND~\citep{ASCEND}       & 10.62 & 23    & Conversation    & $\checkmark$ & $\checkmark$ & $\times$ \\
Ours                        & 104.02   & 200   & Conversation    & $\checkmark$ & $\checkmark$ & $\checkmark$ \\
\bottomrule
\end{tabular}
}
\end{table*}

\section{Related Work}

Existing Mandarin-English CS speech datasets can be broadly categorized into read speech and spontaneous speech corpora. Read speech datasets typically contain pre-defined sentences that participants are instructed to read aloud, offering controlled phonetic and linguistic variations but lacking the spontaneity of natural conversations.  

The CECOS dataset~\citep{shen2011cecos} is one of the earliest Mandarin-English CS corpora, comprising 12.1 hours of read speech from 77 speakers at National Cheng Kung University in Taiwan. While it includes code-switching utterances, it lacks publicly available transcriptions. OC16-CE80~\citep{wang2016oc16} significantly expands the scale, offering 80 hours of read speech from over 1400 speakers, with transcriptions available but not open-sourced. The ASRU dataset~\citep{shi2020asru}, developed for an ASR challenge, contains 240 hours of predominantly Mandarin speech interspersed with some English. Although transcriptions exist, the dataset is not publicly accessible.  

More recent datasets, such as DOTA-ME-CS~\citep{li2025dota}, offer open-source transcriptions and introduce AI-based augmentation techniques (e.g., timbre synthesis, speed variation, and noise addition) to enhance diversity. However, its scale remains relatively small, with only 18.54 hours from 34 speakers. TALCS~\citep{TALCS} provides a much larger dataset, comprising 587 hours of speech from online teaching scenarios. While it is open-source and valuable for acoustic modeling, its domain-specific nature introduces biases in discourse structure, grammar, and lexical choices, making it less representative of everyday spontaneous conversations.  

Spontaneous CS datasets, in contrast, are essential for modeling real-world language use but present greater challenges in collection and annotation. SEAME~\citep{SEAME} provides approximately 30 hours of spontaneous Mandarin-English conversations from 92 speakers in Singapore and Malaysia. It includes word-level transcriptions with time-aligned language boundaries, making it a valuable resource for code-switching research.
\citet{li2012mandarin} compiled 36 hours of spontaneous CS speech across various settings, including conversational meetings and student interviews, but only part-of-speech data is transcribed, limiting its usability for ASR research. ASCEND~\citep{ASCEND} provides a smaller (10.62 hours) yet fully transcribed and open-source dataset of spontaneous CS conversations recorded in Hong Kong, featuring 23 bilingual speakers.  

Despite these advancements, most existing datasets exhibit limitations in scale, availability, or annotation completeness. Many either focus on isolated code-switching utterances rather than full dialogues, or remain inaccessible to the research community. In contrast, our dataset aims to bridge these gaps by providing 104 hours of spontaneous Mandarin-English CS speech, featuring full-length dialogue recordings with comprehensive transcriptions. It captures naturalistic code-switching patterns within extended conversations, making it a valuable resource for both ASR research and broader linguistic analysis.  

While our focus is on Mandarin-English, it is important to acknowledge the growing body of research on code-switching in other language pairs. Notable examples include datasets and studies for Spanish-English~\citep{garcia2018code}, Arabic-English~\citep{chowdhury2021towards}, Hindi-English~\citep{dey2014hindi}, and Manipuri-English~\citep{singh2024mecos}, each contributing to a broader understanding of this complex linguistic phenomenon.

\section{Dataset Creation}

The creation of the CS-Dialogue dataset involved a meticulous multi-stage process, encompassing careful data acquisition and rigorous annotation ensuring the development of a high-quality resource for code-switching research.

\subsection{Data Acquisition}

\subsubsection{Speaker Selection}

All speakers were native Chinese citizens with demonstrated fluency in English.  Selection criteria prioritized individuals with significant exposure to English-speaking environments, such as overseas experience or high scores on standardized English proficiency tests (e.g., IELTS 6 or TEM-4). Prospective speakers underwent an audition to ensure adequate speech quality and language proficiency before being included in the recording sessions.

\subsubsection{Ethical Considerations and Compensation}

Prior to participation, all speakers provided informed consent, granting permission for the collection, processing, and potential sharing of their data, including with parties located outside of China. The consent process adhered to ethical guidelines and ensured participants were fully aware of the data's intended use. Each speaker received financial compensation of 300 RMB (approximately 50 USD ) for their contribution to the dataset. A more detailed discussion of ethical considerations is provided in Appendix~\ref{appendix:ethics}.

\subsubsection{Topic Selection}

The dataset incorporates seven prevalent topics of daily relevance: personal topics, entertainment, technology, education, job, philosophy, and sports. A detailed overview of these topics could be found in Appendix~\ref{appendix: Topics, Descriptions, and Examples}. To ensure comprehensive coverage, a minimum of 15 distinct speaker pairs engaged in discussions for each topic. Individual speaker pairs selected between two and six topics based on their personal interests, aiming to foster natural and engaging conversations.

\subsubsection{Dialogue Recording Procedure}
To facilitate natural and spontaneous interaction, paired dialogues were conducted through an audio-visual platform.
Each participant recorded an individual audio stream using a smartphone microphone in a quiet indoor environment. Smartphone models were not standardized, but the device category (Android or iOS) was retained for each speaker. For privacy and efficiency, only the audio recordings were retained for the dataset, and the released audio is entirely original: no denoising, dereverberation, speech enhancement, or loudness normalization was applied.
Most pairs were unacquainted prior to participation, though a small subset were already acquainted. Regardless of prior familiarity, all sessions followed the same protocol and opened with a brief mutual self-introduction as a natural warm-up.
A timekeeper facilitated each session, ensuring adherence to the established recording protocol. Each dialogue commenced with brief introductory remarks, transitioning into discussions centered on the pre-selected topics. The linguistic composition of the dialogue progressed systematically: initially in Mandarin Chinese, followed by a period of code-switching between Chinese and English, and concluding with exclusive use of English. Each topic segment was designed to last approximately 20 minutes, with a target allocation of 8 minutes for Chinese, 6 minutes for code-switching, and 6 minutes for English. The longer Mandarin phase served as a comfortable warm-up that helped speaker pairs establish the topic before the target mixed phase.

While the timekeeper provided prompts to maintain the intended schedule, natural variations in pacing were permitted to encourage spontaneous and authentic communication. 
Participants were not strictly limited to a single language during any segment. They could code-switch naturally in monolingual phases, and monolingual speech was also allowed during the code-switching segment. This flexible setup helped preserve spontaneity. The transcriptions faithfully reflect what was actually spoken, including deviations from the intended language schedule, ensuring the dataset captures authentic conversational behavior.
A dedicated observer monitored each session, verifying procedural compliance and recording relevant metadata for subsequent analysis.
The entire procedure took approximately 1.5 hours.  All audio files in the dataset are stored in a 16 kHz, 16-bit, mono, PCM WAV format.

\subsection{Annotation}

\begin{wraptable}{r}{8cm}
\centering
\caption{Annotation Symbols and Definitions}
\label{tab:annotation_symbols}
\small
\begin{tabular}{lp{6cm}}
\toprule
Symbol & Definition \\
\midrule
\texttt{**} & Indicates unintelligible words or phrases. \\
\texttt{<FIL/>} & Filled pauses resulting from hesitation. \\
\texttt{<SPK/>} & Speaker-related noises, such as lip smacking, laughter, coughing, or throat clearing. \\
\texttt{<NON/>} & Non-speech noises, such as door slams, knocks, or ringing sounds. \\
\texttt{<NPS/>} & Noises made by individuals other than the designated speakers, including speech or noise. \\
\bottomrule
\end{tabular}

\end{wraptable}

To ensure high data quality and support downstream tasks, all audio files underwent a rigorous annotation process.  This included precise manual transcription, detailed labeling of non-lexical events, and strict quality control procedures. All annotations were carried out by a dedicated in-house team (see Appendix~\ref{appendix_annotator_summary} for annotator details) following a standardized protocol. The data provider used an internally developed annotation tool for transcription and time alignment, with annotations exported in TextGrid format. An illustrative example of a dialogue transcription is provided in Appendix~\ref{appendix_dialogue_case}.
English spoken by our participants reflects typical L2 English pronunciation patterns of Mandarin native speakers. The transcripts are orthographic rather than phonetic: English lexical items use standard English spellings, Chinese speech is represented using Chinese characters, and L2 accents are not transcribed using pronunciation-based spellings. Repetitions, repairs, and disfluencies are preserved as spoken, following the transcription guidelines below.

The transcription process prioritized accurate representation of the spoken content, focusing on the speaker's actual pronunciation. The following guidelines were implemented to maintain consistency and ensure high transcription quality:

\begin{enumerate}
    \item \textbf{Word Count Fidelity:} Transcriptions were required to maintain a precise word-for-word correspondence with the spoken utterance, preventing both omissions and additions.
    \item \textbf{Treatment of Disfluencies:} Clear repetitions of sounds or words were transcribed verbatim (e.g., "\begin{CJK}{UTF8}{gbsn}放放假\end{CJK}" transcribed as "\begin{CJK}{UTF8}{gbsn}放放假\end{CJK}"). Partially articulated syllables were transcribed using the most appropriate homophone (e.g., "\begin{CJK}{UTF8}{gbsn}放假\end{CJK}" pronounced as "fu-fang4-jia4" transcribed as "\begin{CJK}{UTF8}{gbsn}夫放假\end{CJK}"). Epenthetic or extremely faint sounds were disregarded.
    \item \textbf{Numerical Representation:} Arabic numerals were converted to their corresponding Chinese characters or English words, depending on the context and pronunciation (e.g., "711" transcribed as "\begin{CJK}{UTF8}{gbsn}七幺幺\end{CJK}" or "Seven Eleven").
    \item \textbf{Accent Accommodation:} Regional accents and variations in pronunciation (e.g., distinctions between retroflex and non-retroflex consonants, nasal finals, or the pronunciation of /h/ and /f/ or /l/ and /n/) were preserved in the transcription without correction.
    \item \textbf{Punctuation Conventions:} Punctuation marks, including both Chinese and English symbols, were applied according to standard grammatical conventions and semantic context to ensure clarity and accurate segmentation.
    \item  \textbf{Spelling conventions:} Spelling followed common English conventions and standards to ensure quality of annotations.
    \item \textbf{Acronym Representation:} Acronyms were transcribed using uppercase letters separated by spaces (e.g., "I B M"). Utterances consisting of three or fewer letters transcribed as an acronym were categorized as Chinese.
\end{enumerate}

\begin{wraptable}{r}{5.5cm}
\centering
\caption{Overview of our dataset}
\label{tab:overview}
\begin{tabular}{lc}
\toprule
Characteristic      & Value    \\
\midrule
Duration (hrs)      & 104.02   \\
\# Speakers         & 200      \\
\# Dialogues        & 100      \\
\# Raw Recordings   & 200      \\
\# Topic Sessions   & 320      \\
\# Utterances       & 38,917   \\
Avg. Duration (s)   & 9.62     \\
\bottomrule
\end{tabular}
\end{wraptable}
In addition to the transcription of spoken words, a set of specialized symbols was used to annotate non-lexical events and acoustic phenomena. These symbols, detailed in Table \ref{tab:annotation_symbols}, provided additional information about the acoustic characteristics of the data.

Following the initial annotation, a separate quality control team of five members, with no overlap with the annotation team, performed a rigorous two-stage review process: first, a complete review in which every annotation was independently checked by at least one reviewer, followed by an additional sampling-based audit. Identified discrepancies were corrected, and recurring issues were incorporated into revisions of the annotation guidelines, ensuring the high transcription quality.

\section{Dataset Composition and Splits}
This section provides a comprehensive overview of the CS-Dialogue dataset, including its profile, statistical analysis, and details on speaker demographics, duration, topic distribution, and textual characteristics.

\begin{wraptable}{r}{5.5cm}
        \centering
        \small
        \caption{Summary of data splits. The train/dev/test speaker sets (140/30/30) are mutually disjoint, i.e., the split is speaker-independent.}
        \label{tab:data_summary}
        \resizebox{0.85\linewidth}{!}{
        \begin{tabular}{@{}ccccc@{}}
        \toprule
        Split & \# Spk. & \# Utt. & Dur. (hrs) & Avg. (s) \\
        \midrule
        Train & 140     & 26,428  & 68.97      & 9.40     \\
        Dev   & 30      & 6,196   & 18.30      & 10.63    \\
        Test  & 30      & 6,293   & 16.74      & 9.58     \\
        \midrule
        Total & 200     & 38,917  & 104.02     & 9.62     \\
        \bottomrule
        \end{tabular}
        }
\end{wraptable}

\subsection{Profile}

The CS-Dialogue dataset comprises 104.02 hours of spontaneous Mandarin-English code-switching speech from 200 speakers, structured as 100 dialogues (200 raw recordings, as each dialogue involves two participants).  These dialogues encompass 320 topic sessions, offering a diverse range of conversational contexts. The dataset contains 38,917 utterances. Table~\ref{tab:overview} summarizes the key characteristics of the dataset, including the total duration, number of speakers, dialogues, utterances, and language distribution.

To facilitate rigorous model development and evaluation, we partition the dataset into three distinct subsets: training, development, and test. The detailed statistics for each split are summarized in Table~\ref{tab:data_summary}. Crucially, the split is speaker-independent: the 140 training, 30 development, and 30 test speakers are mutually disjoint, ensuring a robust assessment of the model's generalization capabilities to unseen speakers.

\begin{figure*}[!t]
    \centering
    \begin{subfigure}{0.48\linewidth}
        \centering
        \includegraphics[width=\linewidth]{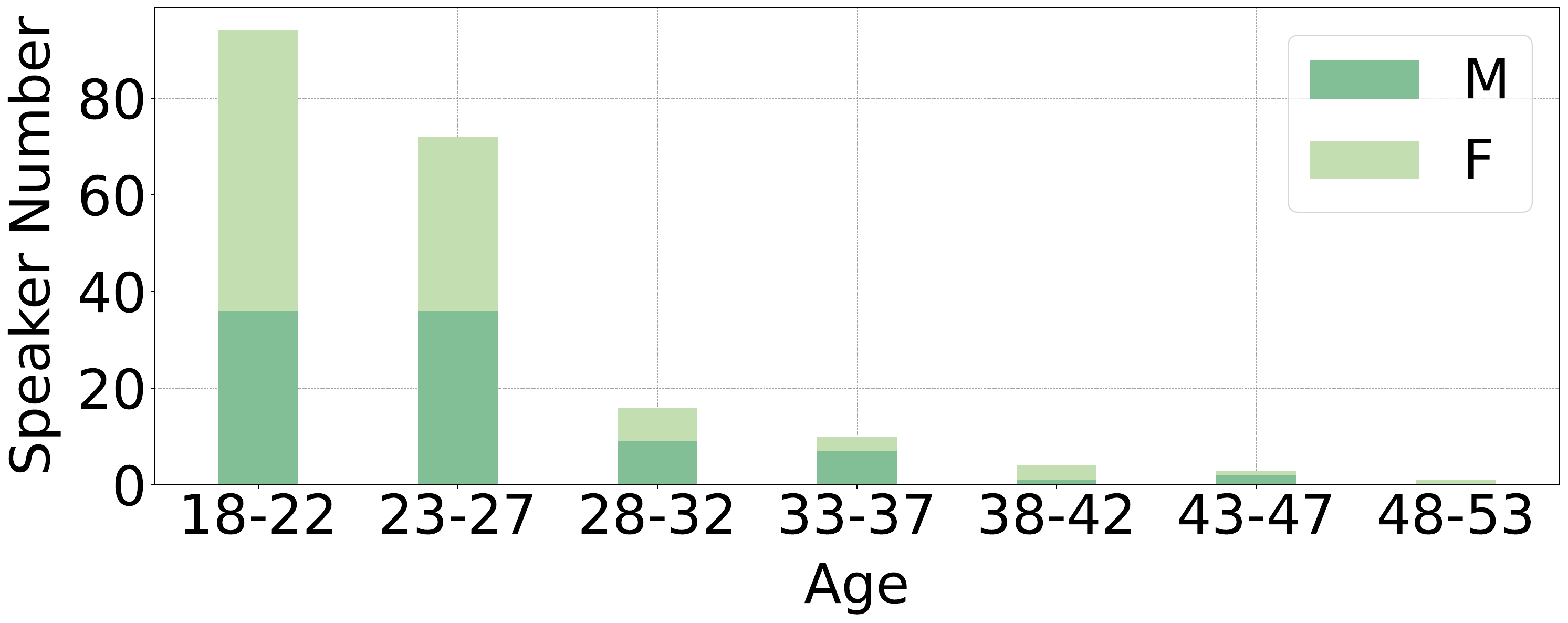}
        \caption{Distribution of speakers by age and gender}
        \label{pic:age_gender}
    \end{subfigure}
    \hfill
    \begin{subfigure}{0.48\linewidth}
        \centering
        \includegraphics[width=\linewidth]{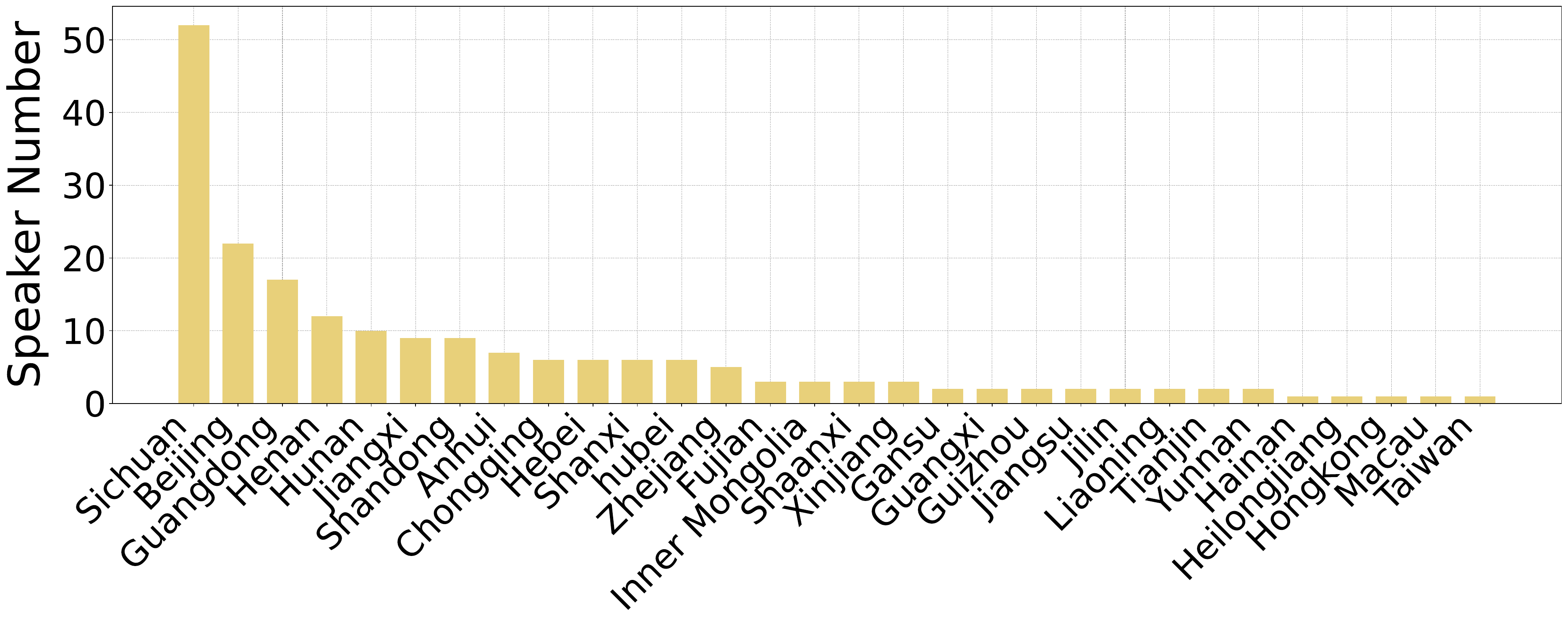}
        \caption{Distribution of speakers by region}
        \label{pic:province}
    \end{subfigure}
    \caption{Demographic distributions of the speaker population}
    \label{pic:speaker_demographics}
\end{figure*}

\subsection{Statistics}

\subsubsection{Speaker Demographics}

The age and gender distribution of the speakers is illustrated in Figure~\ref{pic:age_gender}. Speaker ages range from 18 to 53, grouped into four-year intervals.  Male speakers are represented in green and female speakers in light green in the stacked bar chart.  A notable trend is the concentration of speakers in the younger age brackets (18-22 and 23-27), with a relatively balanced gender distribution.  
\begin{wrapfigure}{r}{6.5cm}
  \centering
  \includegraphics[width=1.0\linewidth]
  {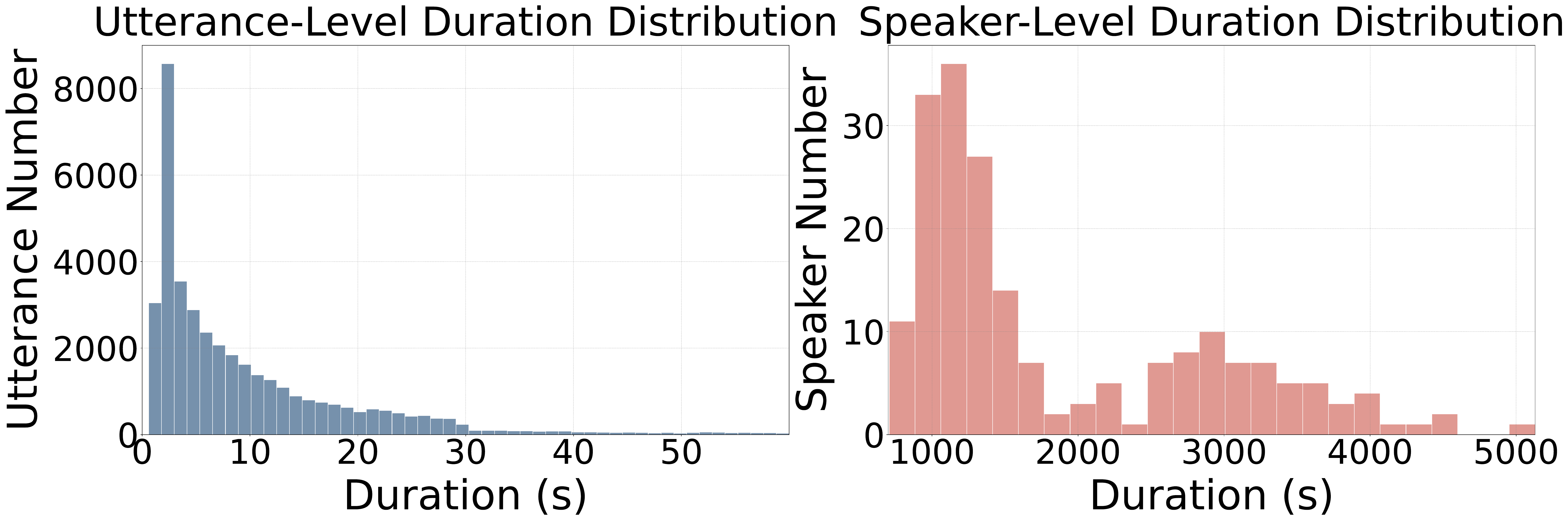}
  \caption{Utterance-level (left) and speaker-level (right) duration distributions.}
  \label{pic:short_duration}
\end{wrapfigure}
The decrease in speaker numbers in older age groups may be attributed to the greater prevalence of Mandarin-English bilingualism among younger generations, or potential challenges in recruiting older participants with the required language proficiency. The data was collected from various regions in China.

The regional distribution of speakers, based on their reported origin, is displayed in Figure~\ref{pic:province}.  Sichuan has the highest representation, followed by Beijing and Guangdong, while the remaining regions have significantly fewer speakers.  A complete tabulated summary of speaker demographics (age, gender, region, and recording device) is provided in Appendix~\ref{appendix:speaker_demographics}.

\subsubsection{Duration Analysis}

\begin{wrapfigure}{r}{5cm}
  \centering
  \includegraphics[width=1.0\linewidth]
  {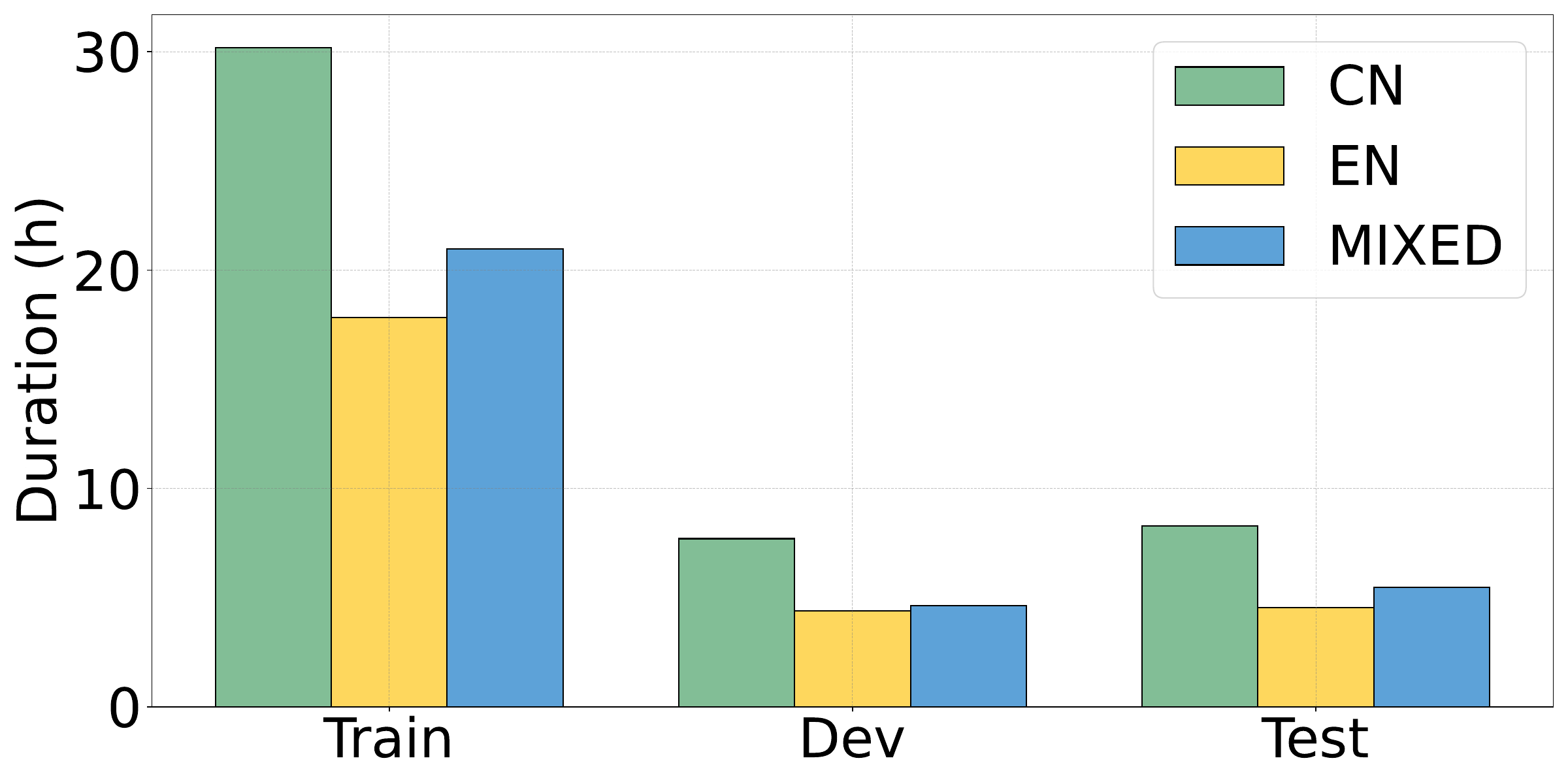}
  \caption{Duration of each language per data split}
  \label{pic:cstype_duration}
\end{wrapfigure}

Utterance-level and speaker-level duration distributions are presented in Figure~\ref{pic:short_duration}.  Most utterances are under 30 seconds, and the majority of speakers have a total speaking time clustered towards the lower end of the range.  However, a few speakers contribute significantly more data, leading to a long-tailed distribution.

The training, development, and test sets exhibit a consistent proportional distribution of Chinese, English, and mixed-language durations, as shown in Figure~\ref{pic:cstype_duration}.  This balanced representation of each language category within each split ensures that models trained on one split are likely to generalize well to others.  Appendix~\ref{appendix: Full-dialogue duration distribution} details the distribution of full-dialogue durations.

\subsubsection{Dialogue Topic Analysis}

The dataset's conversation topics are distributed as shown in Table~\ref{tab:topic_freqency}.  Categorized into seven broad themes—Personal Topics, Entertainment, Technology, Education, Job, Philosophy, and Sports—the 320 topic sessions offer diverse conversational contexts (see Appendix~\ref{appendix: Topics, Descriptions, and Examples} for details). 
Personal Topics are the most frequent (24.38\%; 78/320 sessions), while Philosophy is the least frequent (4.69\%), indicating a focus on everyday conversational themes, along with a smaller, yet still significant, representation of more specialized topics.

\begin{wraptable}{r}{5cm}
\small
\centering
\caption{Topic distribution in the full-dialogue recordings}
\label{tab:topic_freqency}
\resizebox{1.0\linewidth}{!}{
\begin{tabular}{lcc}
\toprule
Topic Name      & Frequency & Proportion \\
\midrule
Personal topics & 78        & 24.38\%        \\
Entertainment   & 60        & 18.75\%        \\
Technology      & 26        & 8.13\%         \\
Education       & 61        & 19.06\%        \\
Job             & 36        & 11.25\%        \\
Philosophy      & 15        & 4.69\%         \\
Sports          & 44        & 13.75\%        \\
Total           & 320       & 100.00\%      \\
\bottomrule
\end{tabular}
}
\end{wraptable}

Further details on the distribution of these seven conversation topics across each data split (training, development, and test sets) are provided in Appendix~\ref{appendix:topic_distribution_details}.

\begin{table*}[t]
\centering
\scriptsize
\caption{Performance of different models training from scratch under various decoding strategies.}
\label{tab:decoding_results}
\resizebox{1.0\linewidth}{!}{
\begin{tabular}{@{}l@{\hspace{7pt}}c@{\hspace{7pt}}c@{\hspace{7pt}}c@{\hspace{7pt}}c@{\hspace{7pt}}c@{\hspace{7pt}}c@{\hspace{7pt}}c@{\hspace{7pt}}c@{\hspace{7pt}}c@{\hspace{7pt}}c@{\hspace{7pt}}c@{\hspace{7pt}}c@{\hspace{7pt}}c@{\hspace{7pt}}c@{\hspace{7pt}}}
\toprule
\multirow{2}{*}{Model} & \multirow{2}{*}{\begin{tabular}[c]{@{}c@{}}\# Params\end{tabular}} & \multicolumn{3}{c}{Greedy} & \multicolumn{3}{c}{Beam} & \multicolumn{3}{c}{Attention} & \multicolumn{3}{c}{\begin{tabular}[c]{@{}c@{}}Attention\\ Rescoring\end{tabular}} \\
\cmidrule(lr){3-5} \cmidrule(lr){6-8} \cmidrule(lr){9-11} \cmidrule(lr){12-14}
                       &          & CER    & WER    & MER    & CER    & WER    & MER    & CER     & WER     & MER     & CER      & WER      & MER      \\
\midrule
Transformer            & 29M      & 22.56  & 45.34  & 27.21  & 22.24  & 45.19  & 27.01  & 39.23  & 62.80  & 44.05  & 21.60    & 43.44    & 26.06   \\
Branchformer           & 29M      & 18.86  & 39.16  & 23.01  & 18.78  & 39.20  & 22.95  & 44.06  & 60.90  & 47.50  & 18.29    & 37.55    & 22.23   \\
Conformer              & 31M      & \textbf{15.91}  & \textbf{33.67}  & \textbf{19.54}  & \textbf{15.88}  & \textbf{33.60}  & \textbf{19.50}  & \textbf{24.98}  & \textbf{42.75}  & \textbf{28.61}  & \textbf{15.45}    & \textbf{32.36}    & \textbf{18.91}  \\
\bottomrule
\end{tabular}
}
\end{table*}

\subsubsection{Text Analysis}

An analysis of frequent strings (Appendix~\ref{appendix: top frequent strings}) reveals distinct patterns in language use and code-switching strategies.  Discourse markers like \begin{CJK}{UTF8}{gbsn}"我觉得"\end{CJK} (I think) and \begin{CJK}{UTF8}{gbsn}"比如说"\end{CJK} (for example) characterize the Chinese segments, while phrases like "a lot of" and "I think it's" are frequently used in the English segments.  A crucial observation in the mixed-language segments is the frequent use of function words from one language to frame content words from the other (e.g., "you know \begin{CJK}{UTF8}{gbsn}就\end{CJK}"), suggesting that code-switching commonly occurs at clause or phrase boundaries.

\section{Experiments}

\subsection{Metrics}
\label{sec:metrics}
ASR performance on the code-switching dataset is evaluated using three metrics: Mixture Error Rate (MER), Word Error Rate (WER), and Character Error Rate (CER).  Following \citep{shi2020asru}, MER is adopted as the primary metric due to its holistic assessment of ASR accuracy.  In MER, Mandarin characters and English words are combined into a shared token sequence, and the edit distance between the reference and hypothesis sequences is computed over this unified tokenization.  WER and CER are calculated separately for English and Chinese segments to provide more granular insights into per-language performance.  The corresponding normalization and scoring scripts are released with the dataset.

\subsection{Baseline Models}
We evaluate two categories of ASR baselines on the CS-Dialogue dataset: models trained from scratch and large-scale pre-trained models. Training and decoding details are provided in Appendix~\ref{Appendix_hyperparem}.
For scratch models, we use WeNet \citep{wenet} to train Transformer \citep{transformer}, Conformer \citep{conformer}, and Branchformer \citep{peng2022branchformer}. All models are trained only on CS-Dialogue with a joint CTC \citep{ctc} and AED \citep{chorowski2014end} loss, without external data.
We also evaluate six pre-trained models: Whisper \citep{radford2023robust}, a multilingual Transformer trained on 680k hours of speech; Qwen2-Audio \citep{chu2024qwen2}, a large-scale audio-language model; SenseVoice-Small \citep{an2024funaudiollm}, a non-autoregressive multilingual speech model; Paraformer \citep{gao22b_interspeech}, a fast non-autoregressive end-to-end ASR model; Voxtral-Mini-4B-Realtime-2602 \citep{voxtral2025}, a multimodal audio understanding model; and Qwen3-Omni-30B-A3B \citep{qwen3omni2026}, a large-scale omni-modal foundation model.

\begin{table*}[!t]
\small
\centering
\caption{Zero-shot performance of pre-trained ASR models on the CS-Dialogue test set. S: Substitution; D: Deletion; I: Insertion.}
\label{tab:asr_results}
\resizebox{0.8\linewidth}{!}{
\begin{tabular}{cccccccc}
\toprule
\multirow{2}{*}{Model} & \multirow{2}{*}{\# Param} & \multirow{2}{*}{CER (\%)} & \multirow{2}{*}{WER (\%)} & \multicolumn{4}{c}{MER (\%)} \\
\cmidrule(lr){5-8}
                       &                           &                           &                           & S     & D     & I     & Overall \\
\midrule
Whisper Large-V2               & 1,550M                    & 10.70                      & 31.11                     & 6.00  & 7.60  & \uline{1.69}  & 15.29   \\
Qwen2-Audio            &   8.2B                 & 7.15                      & \uline{19.82}                     & \uline{4.32}  & 1.82  & 3.62  & 9.76    \\
Paraformer             &  220M                       & \textbf{3.70 }                     & 32.02                     & 6.30  & \uline{0.98}  & 2.37  & \uline{9.65}    \\
SenseVoice-Small       &  234M                         & \uline{4.42}                      & \textbf{15.57}                     & \textbf{3.44}  & 1.42  & 1.85  & \textbf{6.71}    \\
\midrule
Voxtral-Mini-4B-Realtime-2602  & 4B                         & 17.41                      & 39.24                     & 9.01   & 12.36   & \textbf{0.80}   & 22.18   \\
Qwen3-Omni-30B-A3B             & 30B / 3B act.              & 4.80                       & 36.12                     & 7.67   & \textbf{0.74}   & 3.28   & 11.70   \\
\bottomrule
\end{tabular}
}
\end{table*}

\begin{table*}[!t]
\centering
\caption{Zero-shot and fine-tuning performance of different Whisper models and SenseVoice-Small on the CS-Dialogue test set. }
\label{tab:whisper_results}
\resizebox{0.8\linewidth}{!}{
\begin{tabular}{cccccccc}
\toprule
\multirow{2}{*}{Model} & \multirow{2}{*}{\# Param} & \multicolumn{3}{c}{Zero-shot}  & \multicolumn{3}{c}{Fine-tuning} \\
\cmidrule(lr){3-5} \cmidrule(lr){6-8}
                         &                           & CER (\%) & WER (\%) & MER (\%) & CER (\%)  & WER (\%) & MER (\%) \\
\midrule
Whisper-Tiny                     & 38M                       & 27.83    & 41.69    & 31.11    & 19.24     & 29.64    & 21.38    \\
Whisper-Base                     & 74M                       & 19.90    & 37.21    & 23.90    & 15.36     & 27.20    & 17.80    \\
Whisper-Small                    & 244M                      & 12.82    & \uline{30.81}    & 16.76    & 7.51      & 16.09    & 9.26     \\
Whisper-Medium                   & 769M                      & \uline{11.34}    & 32.57    & \uline{15.88}    & \uline{6.12}      & \uline{13.02}    & \uline{7.53}    \\ \midrule
SenseVoice-Small                & 234M                       & \textbf{4.42}     & \textbf{15.57}    & \textbf{6.71}     & \textbf{3.34}      & \textbf{10.87}    & \textbf{4.99}    \\     
\bottomrule
\end{tabular}
}
\end{table*}

\subsection{Main Results}
\label{sec:main_results}

\subsubsection{Performance of Models Trained from Scratch}
The performance comparison of models trained from scratch is presented in Table~\ref{tab:decoding_results}.  Across all decoding methods (greedy decoding, beam search, attention decoding, and attention rescoring), the Conformer consistently outperforms both the Transformer and Branchformer.  Attention rescoring yields the best performance for all models, resulting in the lowest CER, WER, and MER.  For instance, the Conformer achieves a CER of 15.45\%, a WER of 32.36\%, and an MER of 18.91\% with attention rescoring, a substantial improvement over the results obtained with greedy decoding (15.91\% CER, 33.67\% WER, 19.54\% MER).  While the Branchformer generally surpasses the Transformer in performance, it exhibits the highest error rate under the attention decoding strategy.

\subsubsection{Performance of Pre-trained Models}
Table~\ref{tab:asr_results} presents the zero-shot performance of the pre-trained models on the test set. Among them, SenseVoice-Small achieves the lowest MER (6.71\%), while Voxtral-Mini-4B-Realtime-2602 yields the highest (22.18\%).
Despite its broader capabilities and significantly larger size, Qwen2-Audio reports a higher MER (9.76\%) compared to SenseVoice-Small.
Similarly, Whisper Large-V2 and Qwen3-Omni, another two large-scale multilingual models, exhibit MERs of 15.29\% and 11.70\%, respectively.
SenseVoice-Small achieves better MER despite its small size likely because it is optimized for ASR in a limited set of languages, with a task-specific architecture and substantial exposure to Chinese during training. In contrast, larger models like Qwen2-Audio, Whisper, Voxtral-Mini, and Qwen3-Omni are trained for a wide range of tasks and languages, which may dilute their performance on specialized CS-ASR scenarios.
Notably, most models show a higher proportion of substitution errors relative to deletions or insertions, whereas Voxtral-Mini exhibits predominantly deletion errors.

Among the pre-trained models, Whisper is one of the most widely adopted ASR foundation models. We evaluate different sizes of Whisper in both zero-shot and fine-tuned settings, and additionally include SenseVoice-Small, which achieves the best zero-shot performance. The results are presented in Table~\ref{tab:whisper_results}. Fine-tuning consistently yields substantial improvements across all Whisper model sizes. Within the Whisper family, the Medium model achieves the best post-finetuning performance. However, the overall best results are obtained by SenseVoice-Small after fine-tuning, reaching a CER of 3.34\%, a WER of 10.87\%, and an MER of 4.99\%. These findings demonstrate that while larger Whisper models benefit more from fine-tuning, SenseVoice-Small sets the performance benchmark for code-switching ASR in our experiments.

\subsubsection{Error Analysis at Code-Switching Boundaries}
Aggregate error rates alone do not reveal \emph{where} models fail. We therefore conduct a unified boundary analysis for twelve systems on the same 1,041 mixed-language utterances (6,955 boundary tokens and 44,805 non-boundary tokens), where boundary tokens are reference tokens adjacent to a language transition. For all twelve systems, the boundary error rate (BER) is substantially higher than the non-boundary error rate (NBER): for example, Whisper Large-V2 obtains 30.45\% BER versus 13.90\% NBER, and fine-tuned Whisper-Medium still exhibits 13.27\% versus 5.53\%. Substitutions dominate the boundary errors for every model, indicating that models tend to produce an incorrect lexical or language-specific token at a switch rather than simply deleting it. The complete analysis is provided in Appendix~\ref{appendix:boundary_analysis}.

\subsubsection{Impact of Dialogue Context}

\begin{wraptable}{r}{6.5cm}
\centering
\caption{Impact of dialogue context on Whisper Large-V2 performance}
\label{tab:context_results}
\resizebox{1.0\linewidth}{!}{
\begin{tabular}{cccc}
\toprule
Context Segments & CER (\%) & WER (\%) & MER (\%)  \\
\midrule
0 (baseline)               & 10.70 & 31.11 & 15.29 \\
1                & 10.30 & 29.34 & 14.51 \\
2                & 9.81  &28.05  & 13.74 \\
3                & \textbf{9.13}  & \textbf{26.26} & \textbf{12.97} \\
\bottomrule
\end{tabular}
}
\end{wraptable}

To investigate the benefit of utilizing full dialogue context, a characteristic feature of the CS-Dialogue dataset, we conducted an additional experiment with the Whisper Large-V2 model.  Specifically, we evaluated its performance on CS-Dialogue while varying the number of preceding dialogue turns provided as contextual prompts.  For a context size of $k \in \{1,2,3\}$, the reference transcripts of the $k$ immediately preceding segments from the same dialogue were ordered chronologically and concatenated as text context, while only the current audio segment was decoded.  Because ground-truth transcripts rather than predicted ASR outputs were used, this experiment is an oracle-context analysis: it isolates the potential value of dialogue history without introducing error propagation from previous recognition outputs.  The results, presented in Table~\ref{tab:context_results}, demonstrate a clear trend: increasing the amount of dialogue context significantly improves code-switching ASR performance.  For instance, using three preceding dialogue turns as context reduces the MER from 15.29\% (no context) to 12.97\%. This finding highlights the advantage of CS-Dialogue's full dialogue structure over datasets comprising only isolated utterances.

\subsubsection{Performance Analysis}

\paragraph{Cross-corpus evaluations}
We further perform cross-corpus evaluations with SenseVoice-Small and Whisper-Medium on ASCEND, TALCS, and CS-Dialogue. As shown in Table~\ref{tab:cross_corpus}, models fine-tuned on CS-Dialogue not only reduce MER on the in-domain test set, but also consistently outperform their zero-shot counterparts on the two external corpora. This suggests that CS-Dialogue provides useful supervision for learning robust Mandarin--English code-switching patterns that transfer beyond the training corpus.

\begin{wraptable}{r}{6.5cm}
\centering
\caption{Cross-corpus MER (\%) for zero-shot vs.\ fine-tuned models (fine-tuned on CS-Dialogue). Lower is better.}
\label{tab:cross_corpus}
\resizebox{1.0\linewidth}{!}{
\begin{tabular}{lccc}
\toprule
SenseVoice-Small & ASCEND & TALCS & CS-Dialogue \\
\midrule
Zero-shot  & 13.96 & 8.72 & 6.71 \\
Fine-tuned & \textbf{12.22} & \textbf{8.37} & \textbf{4.99} \\
\midrule
Whisper-Medium & ASCEND & TALCS & CS-Dialogue \\
\midrule
Zero-shot  & 24.69 & 19.33 & 15.88 \\
Fine-tuned & \textbf{15.52} & \textbf{16.54} & \textbf{7.53} \\
\bottomrule
\end{tabular}
}
\end{wraptable}

\begin{figure}[!t]
  \centering
  \includegraphics[width=0.6\linewidth]
  {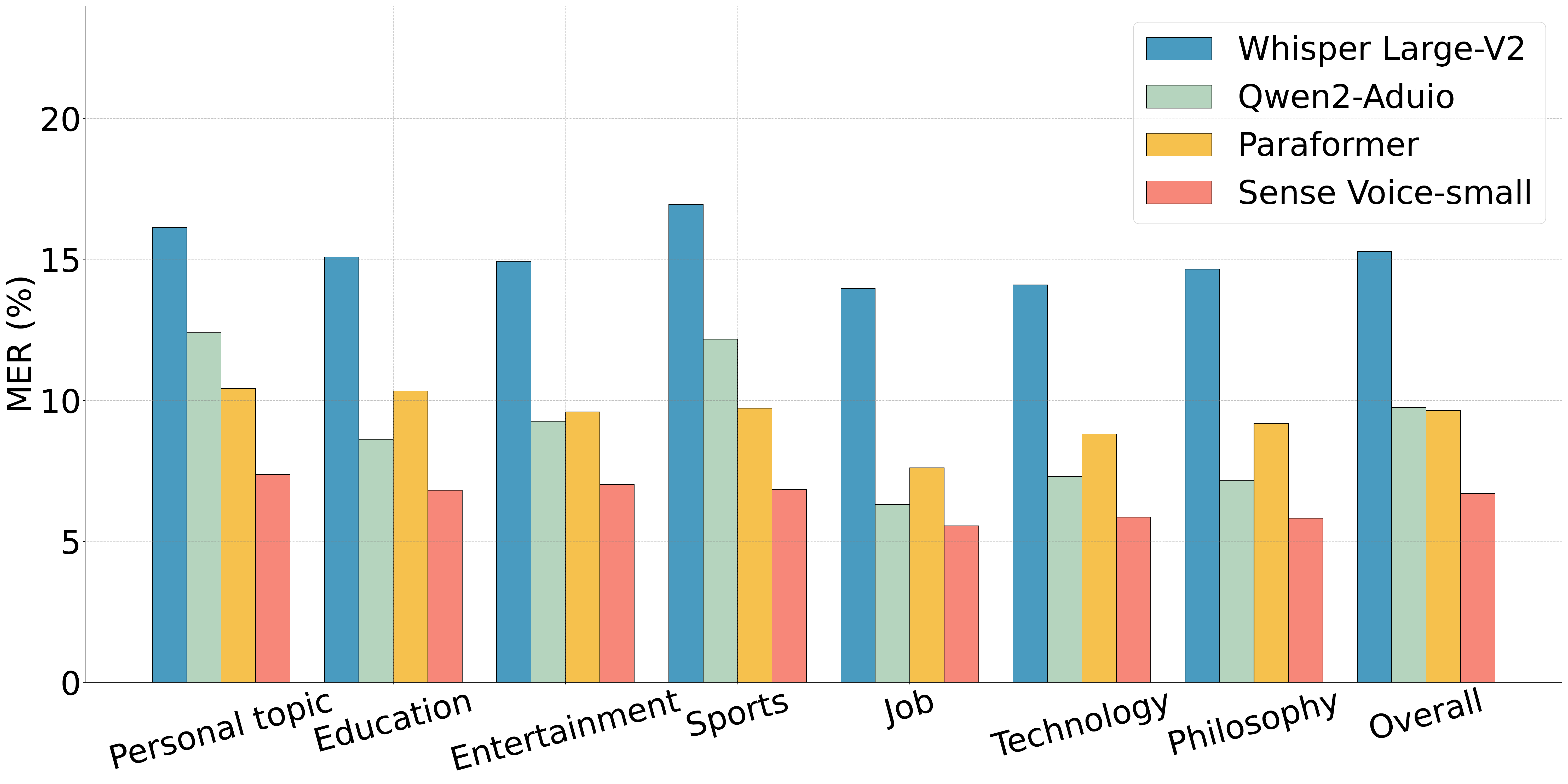}
  \caption{Comparison of MER for Whisper Large-V2, Qwen2-Audio, Paraformer, and SenseVoice-Small across different conversation topics}
  \label{pic:topic_wer}
\end{figure}

\paragraph{Topic-specific Performance Analysis} Figure~\ref{pic:topic_wer} illustrates the MER of the four pre-trained ASR models across the seven conversation topics.  Model performance varies considerably across topics. SenseVoice-Small consistently achieves the lowest MERs, indicating its superior performance on this task.  Comparing Qwen2-Audio and Paraformer reveals no consistent dominance of one model over the other; instead, their relative performance is topic-dependent. In addition, "Sports" and "Philosophy" tend to have higher MERs for all models, while "Job" and "Technology" generally exhibit lower MERs, suggesting varying levels of difficulty across topics.

\paragraph{Real vs. Synthetic Code-Switching Speech} Beyond the main fine-tuning results, we also ask whether synthetic CS speech can replace real bilingual conversations for acoustic model adaptation. As shown in the supplementary experiment (Appendix~\ref{appendix:real_vs_tts_cs}), fine-tuning on real conversational CS audio yields more than a 50\% relative MER reduction compared to TTS-generated CS, indicating that synthetic CS is not a drop-in replacement for real code-switching speech.

\paragraph{Case study} Beyond the quantitative results, a qualitative analysis of the Whisper-Medium model's output is provided in Appendix~\ref{Appendix_case_study}. This analysis includes example transcriptions, comparing zero-shot and fine-tuned performance, and highlights common error types.

\section{Conclusion}
In this paper, we presented CS-Dialogue, a new 104-hour large-scale dataset of spontaneous Mandarin-English code-switching dialogues.  Unlike most existing datasets that primarily offer isolated utterances, CS-Dialogue provides full-length dialogue recordings and complete transcriptions, enabling unprecedented research into the contextual dynamics of code-switching. This resource addresses existing dataset limitations by capturing naturalistic code-switching patterns. Our rigorous data creation and baseline experiments highlight code-switching challenges and the value of fine-tuning. CS-Dialogue offers a benchmark for future ASR and dialogue modeling, aiming to advance robust multilingual communication systems that can leverage conversational context. Future extensions could explore additional language pairs and more diverse conversational settings.

\section*{Acknowledgments}
This work has been supported by the National Key R\&D Program of China (Grant No. 2022ZD0116307) and NSF China (Grant No. 62271270).

\bibliography{example_paper}
\bibliographystyle{icml2026}

\newpage
\appendix
\onecolumn

\section{Dataset details}

\setcounter{figure}{0}  
\setcounter{table}{0}   
\renewcommand{\thefigure}{A.\arabic{figure}}
\renewcommand{\thetable}{A.\arabic{table}}

\subsection{Annotator information} \label{appendix_annotator_summary}
Table~\ref{tab:annotator_summary} provides a breakdown of the annotators' demographic characteristics, including their age, gender, hometown, and educational background.

\subsection{Speaker demographics} \label{appendix:speaker_demographics}
Table~\ref{tab:speaker_demographics} summarizes the available demographic metadata of the 200 speakers, including age, gender, and recording-device category. Participants came from 30 reported regions of origin; the largest groups were Sichuan (52), Beijing (22), Guangdong (17), Henan (12), and Hunan (10), with the remaining regions contributing fewer speakers each. All participants underwent English-proficiency screening (e.g., IELTS 6 or TEM-4) followed by a conversational audition confirming that they could sustain Mandarin--English conversations. Exact per-speaker proficiency scores, overseas-exposure histories, and home-language or dialect backgrounds were not systematically retained during collection (see the Limitations section). Reported regions indicate places of origin and should not be interpreted as native-language, dialect, or English-variant labels.

\begin{table}[htbp]
\caption{Summary of speaker demographics}
\label{tab:speaker_demographics}
\small
\centering
\begin{tabular}{@{}llr@{}}
\toprule
Attribute & Group & Count (Percentage) \\
\midrule
\multirow{6}{*}{Age} & 18--20 & 33 (16.5\%) \\
 & 21--25 & 112 (56.0\%) \\
 & 26--30 & 32 (16.0\%) \\
 & 31--35 & 11 (5.5\%) \\
 & 36--40 & 7 (3.5\%) \\
 & 41--53 & 5 (2.5\%) \\
\midrule
\multirow{2}{*}{Gender} & Female & 109 (54.5\%) \\
 & Male & 91 (45.5\%) \\
\midrule
\multirow{2}{*}{Device} & Android & 102 (51.0\%) \\
 & iOS & 98 (49.0\%) \\
\midrule
Region & 30 reported regions & --- \\
\bottomrule
\end{tabular}
\end{table}

\subsection{Dialogue transcription format} \label{appendix_dialogue_case}

\begin{CJK}{UTF8}{gbsn}
Dialogue transcription format are shown in Figure \ref{appendix_dialogue_case} as an example. Note that he names used in this example (e.g., "凯丽", "贝拉") are pseudonyms and do not correspond to the real names of the speakers, ensuring the privacy of participants.
\end{CJK}

\begin{figure}[h]
  \centering
  \includegraphics[width=0.5\linewidth]
  {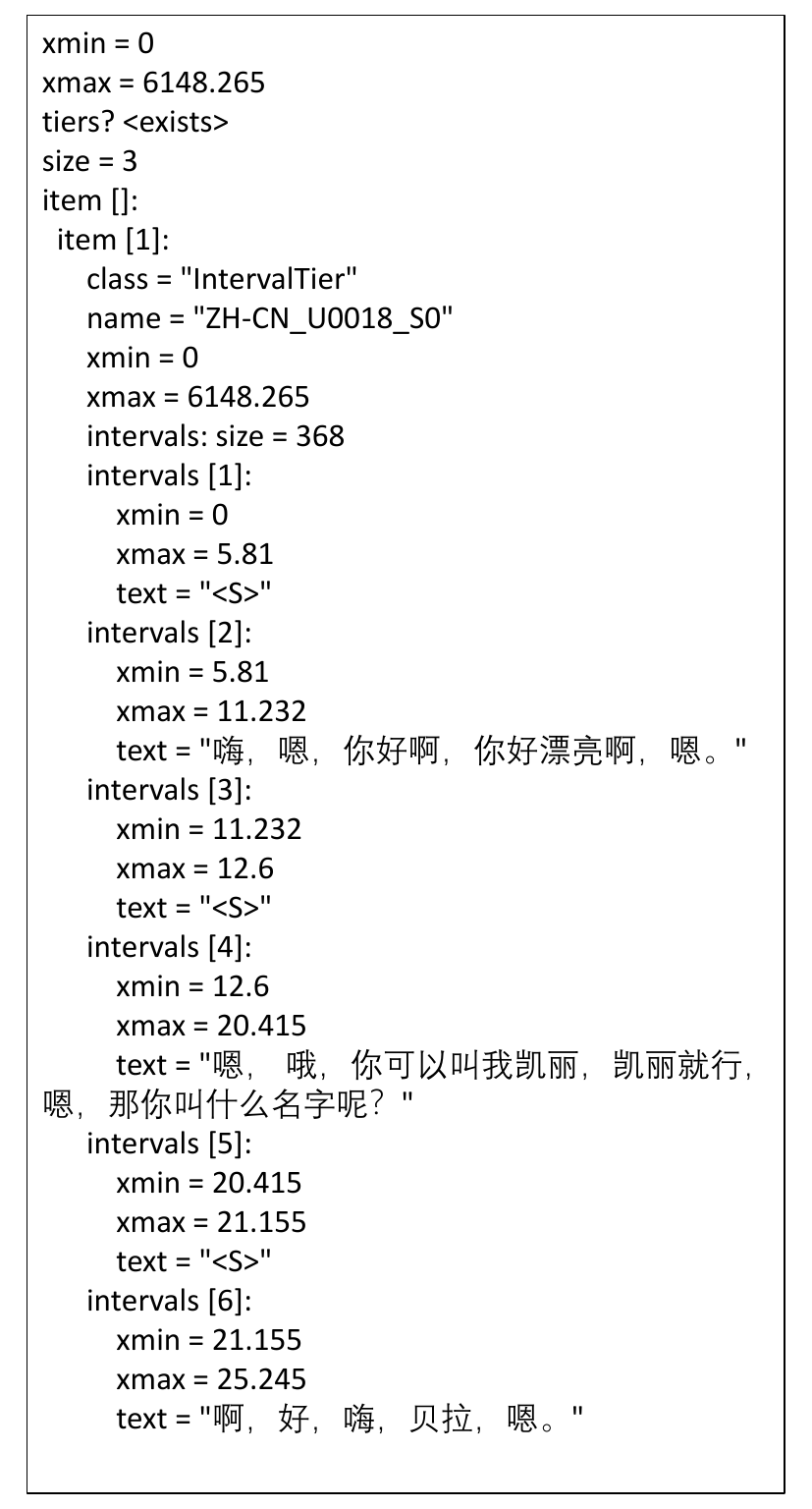}
  \caption{A format example of dialogue transcription file (TextGrid)}
  \label{pic:dialogue_case}
\end{figure}

\begin{table}[htbp]
\caption{Summary of Annotator Demographics}
\label{tab:annotator_summary}
\small
\centering
\begin{tabular}{@{}llr@{}}
\toprule
Category & Value & Count/Percentage \\ \midrule
\multirow{2}{*}{Gender} & Male & 6 (40\%) \\
 & Female & 9 (60\%) \\ \midrule
\multirow{2}{*}{Education} & Bachelor's & 12 (80\%) \\
 & Master's & 3 (20\%) \\ \midrule
\multirow{9}{*}{Hometown} & Liaoning & 1 (6.67\%) \\
 & Henan & 1 (6.67\%) \\
 & Shanxi & 3 (20\%) \\
 & Zhejiang & 1 (6.67\%) \\
 & Jiangxi & 2 (13.33\%) \\
 & Beijing & 1 (6.67\%) \\
 & Fujian & 1 (6.67\%) \\
 & Hebei & 3 (20\%) \\
 & Shaanxi & 1 (6.67\%) \\
& Ningxia & 1(6.67\%) \\
 \midrule
\multirow{5}{*}{Age}
& 21 & 5 (33.33\%)\\
& 22 & 1 (6.67\%)\\
& 24 & 4 (26.67\%) \\
& 25 & 3 (20\%)\\
& 28 & 2 (13.33\%) \\
\bottomrule
\end{tabular}
\end{table}

\begin{table*}[htbp]
\centering
\caption{Details of topics, descriptions, and examples}
\label{tab:topics}
\small
\begin{tabular}{>{\RaggedRight}p{2cm} >{\RaggedRight}p{4cm} >{\RaggedRight}p{6cm}}
\toprule
Topic & Description & Example \\
\midrule
Personal & Discussions centered on individual experiences, preferences, and relationships. &  "\begin{CJK}{UTF8}{gbsn}就是我听你的描述，感觉你喜欢 Taylor. 因为我其实我有个弟弟也很喜欢 Taylor, 就是但是他性格还确实跟你相差蛮大，就是你给我的感觉 You are very a quiet boy\end{CJK}" \\
\midrule
Entertainment & Conversations focusing on various forms of entertainment and cultural trends. & "\begin{CJK}{UTF8}{gbsn}对，因为我们都想表现的自己非常的 courage, 但其实我小的时候也看着也非常 frighten，然后我会直接<FIL/>放学到 home 之后就一直坐坐在 sofa 上面看到十点都 can't move\end{CJK}" \\
\midrule
Technology & Debates and dialogues concerning technological advancements and their impact. & "\begin{CJK}{UTF8}{gbsn}是的，而且他们会通过算法去非常精准地知道你到底想要看一些什么样的东西，所以我感觉其实有的时候 big data 也是一个非常恐怖的东西\end{CJK}" \\
\midrule
Education & Discussions about the academic environment, including challenges and experiences. & "\begin{CJK}{UTF8}{gbsn}那是你们这个 group 自己去想一个 topic 呢，还是这个 professor 会提供一些他的 project 来支撑你们的毕业论文\end{CJK}" \\
\midrule
Job & Conversations about past or present employment situations, work environment, co-workers, etc. &  "\begin{CJK}{UTF8}{gbsn}对对，是如果有有更多的 opportunity 去供我去选择的话，我还是可能会就是放开专业去选择更多的这个看一看，开阔一下我的 horizon\end{CJK}"\\
\midrule
Sports & Discussions on sports activities, athletes, benefits of exercise, etc. & "\begin{CJK}{UTF8}{gbsn}但是这种持续的时间 I couldn't find very long， 就是我也能找到那种 feeling， but very quickly， It disappear only maybe half an hour is a longest period， sometimes 也就十五分钟二十分钟\end{CJK}"\\
\midrule
Philosophy &  Discussions about philosophical ideas and debates on current social issues. & "\begin{CJK}{UTF8}{gbsn}这就是他们现在所处的一个 dilema，It's really classic， I feel like it's happening everywhere。 因为每一个国家都有他各自的 minorities，然后也不得不承认有些地方的 educational resource 真的没有另一些地方更加的 advanced，更加的丰富。\end{CJK}"\\
\bottomrule
\end{tabular}
\end{table*}

\subsection{Topics, descriptions, and examples} \label{appendix: Topics, Descriptions, and Examples} 
Table~\ref{tab:topics} provides details on the seven conversation topics covered in the dataset, including a brief description of each topic and an example utterance illustrating typical content and code-switching patterns. This information clarifies the thematic scope of the data and provides context for interpreting the experimental results.

\section{Dataset analysis} 

\setcounter{figure}{0}  
\setcounter{table}{0}   
\renewcommand{\thefigure}{B.\arabic{figure}}
\renewcommand{\thetable}{B.\arabic{table}}

\subsection{Full-dialogue duration distribution} 
\label{appendix: Full-dialogue duration distribution} 
\begin{figure}[h]
  \centering
  \includegraphics[width=0.45\linewidth]
  {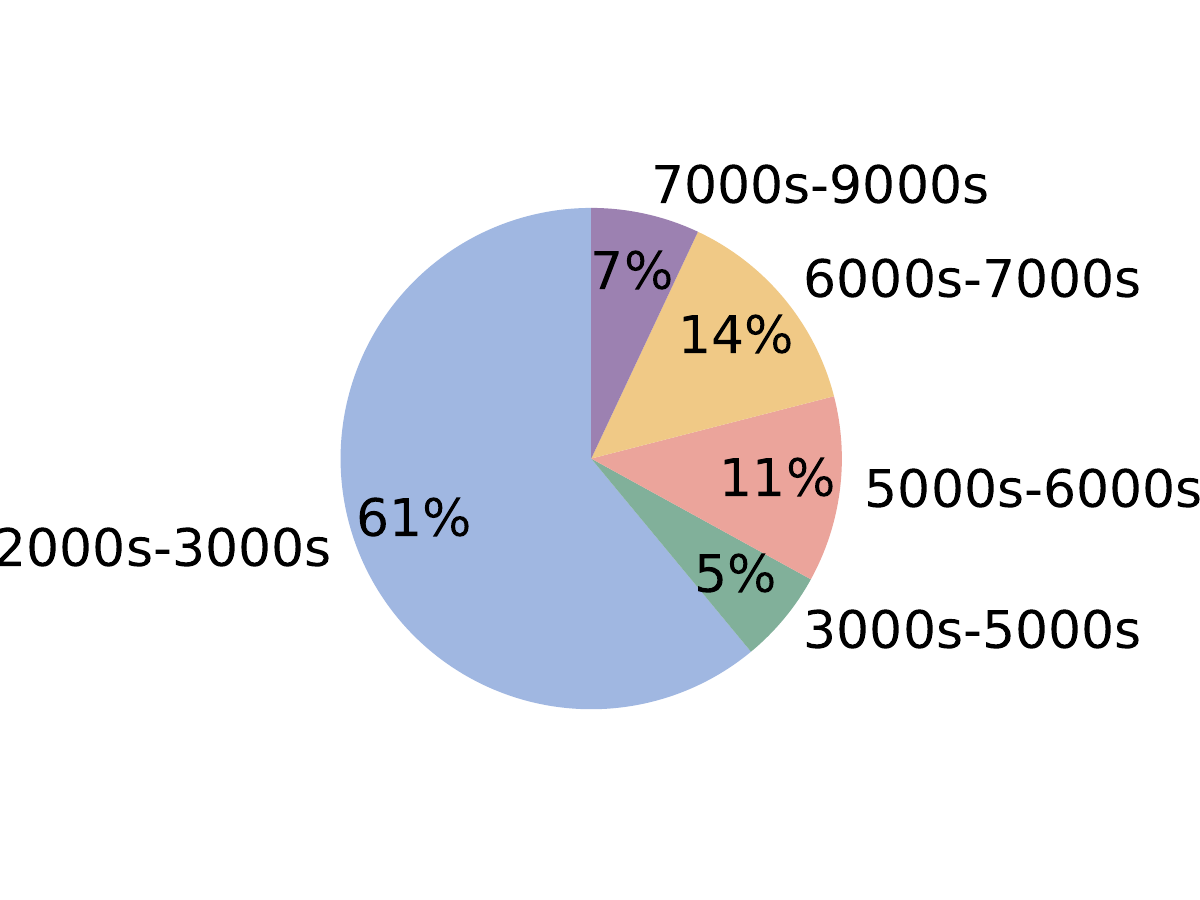}
  \caption{Duration distribution of full-dialogue recordings}
  \label{pic:long_duration}
\end{figure}

As presented in Figure~\ref{pic:long_duration}, most of full-dialogue recordings are between 2000 to 3000 seconds. This distribution indicates a dataset primarily composed of relatively shorter full-dialogue recordings, with a smaller number of significantly longer recordings.

\subsection{Code-switching data distribution}

\begin{table}[htbp]
\centering
\caption{MIXED Statistics for Train/Dev/Test Splits}
\label{tab:mixed_splits_stats}
\begin{tabular}{cccccccc}
\toprule
\multirow{2}{*}{Split} & \multicolumn{3}{c}{Ratio (\%)} & \multicolumn{4}{c}{Duration (h)} \\
\cline{2-8}
 & CN & EN & MIX & CN & EN & MIX & Total \\
\midrule
Train & 43.76 & 25.86 & 30.38 & 30.18 & 17.84 & 20.95 & 68.97 \\
Dev   & 45.26 & 24.89 & 29.85 & 8.28  & 4.55  & 5.46  & 18.30 \\
Test  & 46.00 & 26.26 & 27.74 & 7.70  & 4.40  & 4.64  & 16.74 \\
\bottomrule
\end{tabular}
\end{table}

As shown in Table~\ref{tab:mixed_splits_stats}, the three splits exhibit highly consistent mixing ratios, confirming that the development and test sets are representative of the training distribution.

\subsection{Topic distribution}
\label{appendix:topic_distribution_details}
The distribution of the seven conversation topics across each data split is detailed in Table~\ref{tab:topic_distribution}.  This table presents, for each topic, the total duration, its proportion of the entire dataset, the utterance count, total duration, and average utterance length for each split. While the topic distribution is relatively consistent across the three sets, some variation exists in average utterance lengths. Notably, "Philosophy" tends to have slightly longer utterances than other categories, particularly in the test set (12.95s).

\begin{table*}[htbp]
\centering
\scriptsize
\caption{Topic distribution across training, development, and test sets: counts, durations, and average utterance lengths}
\label{tab:topic_distribution}
\resizebox{1.0\linewidth}{!}{
\begin{tabular}{@{}l@{\hspace{6pt}}c@{\hspace{6pt}}c@{\hspace{6pt}}c@{\hspace{6pt}}c@{\hspace{6pt}}c@{\hspace{6pt}}c@{\hspace{6pt}}c@{\hspace{6pt}}c@{\hspace{6pt}}c@{\hspace{6pt}}c@{\hspace{6pt}}c@{}}
\toprule
\multirow{2}{*}{Topic} & \multirow{2}{*}{\begin{tabular}[c]{@{}c@{}}Dur. (hrs)\end{tabular}} & \multirow{2}{*}{\begin{tabular}[c]{@{}c@{}}Proportion\end{tabular}} & \multicolumn{3}{c}{Train} & \multicolumn{3}{c}{Dev} & \multicolumn{3}{c}{Test} \\
\cmidrule(lr){4-6} \cmidrule(lr){7-9} \cmidrule(lr){10-12}
 &  &  & Count & \begin{tabular}[c]{@{}c@{}}Dur. (hrs)\end{tabular} & \begin{tabular}[c]{@{}c@{}}Avg. (s)\end{tabular} & Count & \begin{tabular}[c]{@{}c@{}}Dur. (hrs)\end{tabular} & \begin{tabular}[c]{@{}c@{}}Avg. (s)\end{tabular} & Count & \begin{tabular}[c]{@{}c@{}}Dur. (hrs)\end{tabular} & \begin{tabular}[c]{@{}c@{}}Avg. (s)\end{tabular} \\ \midrule
Personal   topics      & 21.53                       & 20.69\%                     & 5,865  & 14.59      & 8.95     & 1,348  & 3.55       & 9.48     & 1,487  & 3.39       & 8.22     \\
Education              & 19.20                        & 18.45\%                     & 4,853  & 13.34      & 9.9      & 1,014  & 3.11       & 11.04    & 931   & 2.75       & 10.63    \\
Entertainment          & 23.88                       & 22.95\%                     & 6,609  & 15.85      & 8.63     & 1,509  & 4.39       & 10.47    & 1,402  & 3.64       & 9.34     \\
Sports                 & 14.14                       & 13.59\%                     & 3,476  & 8.64       & 8.95     & 745   & 2.35       & 11.34    & 1,236  & 3.15       & 9.17     \\
Job                    & 11.94                       & 11.48\%                     & 2,555  & 7.37       & 10.38    & 1,052  & 3.02       & 10.35    & 513   & 1.55       & 10.9     \\
Technology             & 8.70                         & 8.36\%                      & 2,151  & 6.09       & 10.19    & 335   & 1.24       & 13.36    & 475   & 1.37       & 10.35    \\
Philosophy             & 4.65                        & 4.47\%                      & 919   & 3.11       & 12.17    & 193   & 0.64       & 11.92    & 249   & 0.9        & 12.95 \\
\bottomrule
\end{tabular}
}
\end{table*}

We further analysis distribution across topic categories in Tabel~\ref{tab:topic_distribution_appendix}. These topic-level statistics indicate that MIXED usage varies meaningfully across domains, with technology-related discussions exhibiting the highest degree of code-switching.
\begin{table}[htbp]
\centering
\caption{Distribution across Topic Categories}
\label{tab:topic_distribution_appendix}
\begin{tabular}{lrrrr}
\toprule
Topic & CN (\%) & EN (\%) & MIXED (\%) & Duration (h) \\
\midrule
Personal      & 46.97 & 25.55 & 27.48 & 21.53 \\
Entertainment & 43.42 & 26.33 & 30.25 & 23.86 \\
Technology    & 35.33 & 23.67 & 41.00 & 8.70  \\
Education     & 43.78 & 26.47 & 29.75 & 19.20 \\
Job           & 45.51 & 24.41 & 30.08 & 11.94 \\
Philosophy    & 42.45 & 27.99 & 29.56 & 4.64  \\
Sports        & 48.15 & 25.80 & 26.05 & 14.14 \\
\bottomrule
\end{tabular}
\end{table}

\section{Text analysis}

\setcounter{figure}{0}  
\setcounter{table}{0}   
\renewcommand{\thefigure}{C.\arabic{figure}}
\renewcommand{\thetable}{C.\arabic{table}}

\label{appendix: top frequent strings} 
\paragraph{Top frequent strings.} To illustrate common linguistic patterns and code-switching behaviors, Table~\ref{tab:frequent_strings} presents the most frequent strings found in the Chinese (CN), English (EN), and mixed-language utterances within the dataset. The table lists the strings and their corresponding frequencies.

\begin{table*}[htbp]
\centering
\small
\caption{Top Frequent Strings in Each Language Category}
\label{tab:frequent_strings}
\begin{CJK}{UTF8}{gbsn} 
\begin{tabular}{cccccc}
\toprule
\multicolumn{2}{c}{CN} & \multicolumn{2}{c}{EN} & \multicolumn{2}{c}{MIXED} \\
\cmidrule(lr){1-2} \cmidrule(lr){3-4} \cmidrule(lr){5-6}
String     & Count     & String         & Count & String           & Count  \\
\midrule
我觉得        & 3,391      &  a lot of      & 304   & you know 就       & 20     \\
的时候        & 2,056      & I think it's   & 208   & know 就是          & 18     \\
比如说        & 1,410      & I want to      & 182   & 这个 AI            & 14     \\
是一个        & 1,173      & do you like    & 178   & school 的时        & 12     \\
因为我        & 1,119      & do you have    & 168   & 我的 friends       & 11     \\
然后我        & 1,033      & I don't know   & 143   & 就是 you           & 11     \\
的一个        & 982       & yeah yeah yeah  & 142   & 就是 AI            & 10     \\
但是我        & 958       & and I think   & 141   & 一个 very          & 10     \\
的一些        & 925       & so I think     & 139   & 非常 interesting   & 10     \\
就是我        & 913       & yeah I think   & 132   & I think 我        & 9      \\
的就是        & 809       & what do you    & 118   & 常的 happy         & 9      \\
非常的        & 792       & go to the      & 115   & play 麻将          & 9      \\
或者是        & 733       & but I think    & 110   & high school 的    & 8      \\
有一些        & 730       & I think I      & 108   & 一个 big           & 8      \\
我感觉        & 726       & you want to    & 107   & together 然后      & 7      \\
\bottomrule
\end{tabular}
\end{CJK} 
\end{table*}

\paragraph{Phoneme-level speaking rate.}
For a more linguistically grounded cross-lingual comparison, we report speaking rate in phonemes per second for Mandarin-only (CN), English-only (EN), and code-switching (MIXED) segments (Table~\ref{tab:pps}). The estimated rates are similar across conditions, with English slightly lower than Mandarin and MIX remaining comparable, suggesting stable speaking tempo during code-switching.

\begin{table}[htbp]
\centering
\caption{Speaking rate (phonemes per second) across segments.}
\label{tab:pps}
\begin{tabular}{lccc}
\toprule
Segment & CN & EN & MIXED \\
\midrule
Phonemes per second & 6.297 & 5.702 & 6.366 \\
\bottomrule
\end{tabular}
\end{table}
\paragraph{English phoneme ratio per utterance}
Table~\ref{tab:eng_phoneme_ratio_mixed} shows that MIXED utterances span a broad spectrum of English usage, with a large concentration in the mid-range (10--40\%), which is characteristic of natural code-switching.
Some MIXED utterances appear to have very high English proportions because MIXED is defined at the utterance level.
In spontaneous dialogue, speakers often switch in bursts (for technical terms, narrative emphasis, or repairs), resulting in locally English-dominant segments while the surrounding discourse remains bilingual.

\begin{table}[htbp]
\centering
\caption{Distribution of English Phoneme Ratio in MIXED Utterances}
\label{tab:eng_phoneme_ratio_mixed}
\begin{tabular}{lrr}
\toprule
English Phoneme Ratio & Count & Percentage \\
\midrule
$[0, 0.1)$   & 1090 & 15.22\% \\
$[0.1, 0.2)$ & 1558 & 21.77\% \\
$[0.2, 0.3)$ & 1421 & 19.86\% \\
$[0.3, 0.4)$ & 1052 & 14.70\% \\
$[0.4, 0.5)$ & 631  & 8.82\%  \\
$[0.5, 0.6)$ & 428  & 5.98\%  \\
$[0.6, 0.7)$ & 283  & 3.95\%  \\
$[0.7, 0.8)$ & 225  & 3.14\%  \\
$[0.8, 0.9)$ & 227  & 3.17\%  \\
$[0.9, 1.0]$ & 244  & 3.41\%  \\
\bottomrule
\end{tabular}
\end{table}

\section{Experimental configurations} \label{Appendix_hyperparem}

\setcounter{figure}{0}  
\setcounter{table}{0}   
\renewcommand{\thefigure}{D.\arabic{figure}}
\renewcommand{\thetable}{D.\arabic{table}}

This section provides detailed hyperparameters used for training and fine-tuning ASR models discussed in the paper. All experiments were conducted using four GTX 3090 or GTX 4090 for several hours. All models utilized in this research are open-source and operate under the MIT License.

\subsection{Training ASR model from scratch}
Table \ref{tab:scratch setting} presents the training hyperparameters for Transformer, Branchformer and Conformer using Wenet toolkit, including batch size, learning rate and epochs. A dynamic batch size is utilized, constrained to a maximum of 60,000 frames per batch.
\begin{table}[htbp]
\centering
\small
\caption{Hyperparameters for training ASR models from scratch.}
\label{tab:scratch setting}
\begin{tabular}{ccccc}
\toprule
\textbf{Model}    & \textbf{Batch size} & \textbf{Learning rate} & \textbf{Epochs} \\ 
\midrule
Transformer    & Dynamic         & 1.00E-03            & 150              \\
Branchformer   & Dynamic         & 1.00E-03              & 150              \\
Conformer     & Dynamic         & 1.00E-03              & 150              \\
\bottomrule         
\end{tabular}
\end{table}

\subsection{Fine-tuning ASR Model}
Table~\ref{tab:finetuning setting} presents the hyperparameters used during fine-tuning of the different Whisper model versions and SenseVoice-Small.  These parameters include the learning rate and the number of epochs. A dynamic batch size is utilized, constrained to a maximum of 12,000 frames per batch.

\begin{table}[htbp]
\centering
\small
\caption{Hyperparameters for fine-tuning different Whisper versions and SenseVoice-Small.}
\label{tab:finetuning setting}
\begin{tabular}{cccc}
\toprule
\textbf{Model}           & \textbf{Batch size} & \textbf{Learning rate}         & \textbf{Epochs} \\
\midrule
Whisper-Tiny              & 16         & 1.00E-05              & 20              \\
Whisper-Base                &16         & 1.00E-05                  & 20               \\
Whisper-Small              & 16         & 1.00E-05                 & 20               \\
Whisper-Medium           & 16         & 1.00E-05                 & 20               \\ \midrule
SenseVoice-Small        &Dynamic     &4.00E-05                  & 10 \\
\bottomrule
\end{tabular}
\end{table}

\section{Comparison with Synthetic Code-Switching Speech Data}
\label{appendix:real_vs_tts_cs}

\setcounter{figure}{0}  
\setcounter{table}{0}   
\renewcommand{\thefigure}{E.\arabic{figure}}
\renewcommand{\thetable}{E.\arabic{table}}

To assess whether synthetic CS speech can substitute real bilingual conversations, we compare fine-tuning on real versus TTS-generated CS audio. Synthetic utterances are produced by CosyVoice2~\cite{du2024cosyvoice} using text from the CS-Dialogue training split, and both real and synthetic sets are used to fine-tune the same Conformer model under identical settings.

\begin{table}[h]
    \centering
    \caption{Conformer performance on CS-Dialogue when fine-tuned on real versus TTS-generated
    code-switching (CS) speech. Lower is better.}
    \label{tab:real_vs_tts_cs}
    \begin{tabular}{llccc}
        \toprule
        \textbf{Training data} & \textbf{Decoding} & \textbf{CER} & \textbf{WER} & \textbf{MER} \\
        \midrule
        Synthetic CS & Greedy           & 32.98 & 59.42 & 38.57 \\
        Real CS      & Greedy           & 15.91 & 33.67 & 19.54 \\
        Synthetic CS & Attn.\ rescoring & 32.50 & 58.11 & 37.92 \\
        Real CS      & Attn.\ rescoring & 15.45 & 32.36 & 18.91 \\
        \bottomrule
    \end{tabular}
\end{table}

Fine-tuning on real conversational CS speech yields substantially better results, with over a 50\% relative reduction in MER compared to synthetic CS data under both decoding strategies. This suggests that TTS-generated CS, even when text-matched, does not fully capture the prosodic patterns, hesitations, and topic-driven switching behavior present in natural bilingual interaction. The observed gap may also stem from the absence of topic-conditioned lexical choices, dialogue and discourse context, natural phrase- and clause-level switch locations, and spontaneous repairs and disfluencies in TTS-generated data.

\section{Case Studies} \label{Appendix_case_study}

\setcounter{figure}{0}  
\setcounter{table}{0}   
\renewcommand{\thefigure}{F.\arabic{figure}}
\renewcommand{\thetable}{F.\arabic{table}}

\begin{figure*}[h]
  \centering
  \includegraphics[width=1.0\linewidth]
  {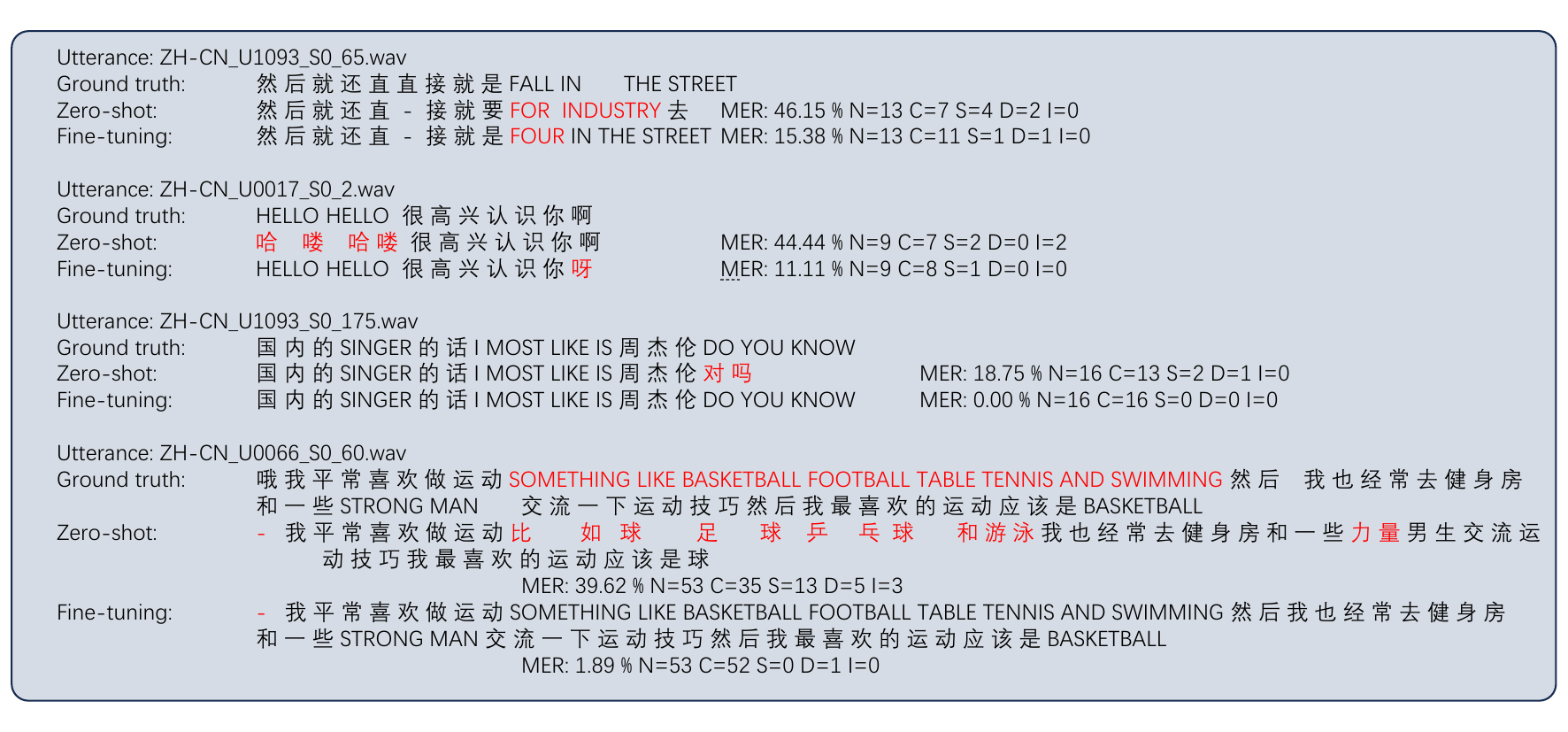}
  \caption{Examples of ASR output from the Whisper Medium model under zero-shot and fine-tuned conditions, showing ground truth transcriptions and error rates}
  \label{pic:some_case}
\end{figure*}

To illustrate the types of errors made by the Whisper Medium model and the improvements achieved through fine-tuning, Figure~\ref{pic:some_case} presents example transcriptions for several utterances. The figure compares the zero-shot and fine-tuned outputs against the ground truth transcriptions, highlighting differences and providing the associated MER. We obsever that whisper designed for both ASR and S2TT tasks, exhibits an unintended behavior in code-switching ASR scenarios. Specifically, the model occasionally produces translations of the input speech rather than accurate transcriptions, deviating from the expected ASR output.

We further observed similar translation-like behavior in the Qwen3-Omni-30B-A3B model. For example, for the utterance \texttt{ZH-CN\_U0066\_S0\_248.wav}, the reference transcription contains ``OH YES AND MY FAVOURITE COMEDY FILM ... \begin{CJK}{UTF8}{gbsn}周星驰\end{CJK}'S FILMS ...'', whereas the model outputs ``OH YES, AND MY FAVORITE COMEDY FILM ... STEPHEN CHOW'S FILMS ...''. Although semantically correct, the model replaces the original Chinese entity with its English form rather than faithfully transcribing the spoken language. We also observed a converse case in which English speech was partially rendered in Chinese. These cases suggest a tension between multilingual semantic generation and language-faithful transcription.

\section{Boundary Error Analysis}
\label{appendix:boundary_analysis}

\setcounter{figure}{0}  
\setcounter{table}{0}   
\renewcommand{\thefigure}{G.\arabic{figure}}
\renewcommand{\thetable}{G.\arabic{table}}

\paragraph{Setup.}
The analysis is conducted on the same 1,041 mixed-language utterances of the test split for all twelve model configurations, containing 6,955 boundary tokens and 44,805 non-boundary tokens. Following the MER tokenization, each Mandarin character and each English word counts as one evaluation token. \emph{Boundary tokens} are reference tokens adjacent to a language transition (a switch between Mandarin and English within the token sequence). We define the boundary and non-boundary error rates as
\[
\mathrm{BER} = \frac{S_B + D_B}{N_B}, \qquad
\mathrm{NBER} = \frac{S_{NB} + D_{NB}}{N_{NB}},
\]
where $S_B$/$D_B$ and $S_{NB}$/$D_{NB}$ denote substitutions and deletions on boundary and non-boundary tokens, and $N_B = 6{,}955$, $N_{NB} = 44{,}805$. Insertions are excluded from both rates because an insertion cannot be uniquely assigned to a reference token.

\begin{table}[htbp]
\centering
\caption{Boundary vs.\ non-boundary error rates for twelve systems on the 1,041 mixed-language test utterances. B-S/B-D: substitutions/deletions on boundary tokens (\%); BER/NBER: boundary/non-boundary error rates (\%); Rel.\ increase: relative increase of BER over NBER.}
\label{tab:boundary_analysis}
\small
\begin{tabular}{lrrrrr}
\toprule
Model & B-S & B-D & BER & NBER & Rel.\ increase \\
\midrule
Whisper-Tiny         & 42.76 & 12.06 & 54.82 & 29.77 & 84.16\% \\
Whisper-Base         & 31.37 & 12.03 & 43.41 & 21.66 & 100.40\% \\
Whisper-Small        & 20.14 & 11.43 & 31.57 & 14.66 & 115.33\% \\
Whisper-Medium       & 22.06 & 12.55 & 34.61 & 15.15 & 128.40\% \\
Whisper Large-V2     & 17.66 & 12.80 & 30.45 & 13.90 & 119.15\% \\
FT Whisper-Tiny      & 25.00 & 4.73  & 29.73 & 15.84 & 87.74\% \\
FT Whisper-Base      & 18.69 & 4.31  & 23.00 & 11.51 & 99.87\% \\
FT Whisper-Small     & 12.24 & 3.18  & 15.41 & 6.98  & 120.85\% \\
FT Whisper-Medium    & 9.83  & 3.44  & 13.27 & 5.53  & 139.86\% \\
Paraformer           & 19.04 & 2.19  & 21.22 & 4.22  & 403.37\% \\
Qwen2-Audio          & 15.74 & 4.64  & 20.39 & 5.33  & 282.37\% \\
SenseVoice-Small     & 9.68  & 3.49  & \textbf{13.17} & \textbf{3.87} & 239.92\% \\
\bottomrule
\end{tabular}
\end{table}

\paragraph{Results.}
Table~\ref{tab:boundary_analysis} reports the results. The boundary error rate is higher than the non-boundary rate for all twelve systems; for example, Whisper Large-V2 obtains 30.45\% BER versus 13.90\% NBER. Fine-tuning substantially improves Whisper-Medium, but boundary tokens remain markedly more difficult (13.27\% versus 5.53\%). SenseVoice-Small achieves the lowest BER, yet still exhibits a large boundary/non-boundary gap. Substitutions dominate the boundary S+D errors for every model; for Paraformer they account for 89.70\%. Thus, models more often produce an incorrect lexical or language-specific token at a switch than simply delete the token, confirming that code-switching boundaries are a systematic difficulty even for strong models.

\section{Ethics Statement}
\label{appendix:ethics}
The collection and use of the CS-Dialogue dataset were conducted in accordance with established ethical guidelines and regulations for human subjects research. Prior to participation, all speakers were provided with a comprehensive information sheet detailing the study's purpose, data collection procedures, and their rights as participants, including the right to withdraw from the study at any time without penalty. Informed consent was obtained from each speaker, explicitly authorizing the recording of their conversations, the processing and analysis of their speech data, and the potential sharing of anonymized data with other researchers (including those located outside of China) for research purposes.

Participants were assured that their data would be treated with strict confidentiality and anonymized to protect their privacy. No personally identifiable information (e.g., names, residential address) will be included in the released dataset or any associated publications. Any segments containing such personally identifiable information will be excluded and treated as invalid.
Participants received compensation for their time and contribution to the study, commensurate with standard rates for similar research participation.  The research protocol, including the informed consent process and compensation procedures, was designed to ensure the protection of participants' rights and well-being. To mitigate potential risks, the topics of discussion during the dialogues were carefully selected to avoid sensitive or potentially harmful content. Participants were given the autonomy to choose topics from a predefined list and were free to pause or stop the recording at any point during the session.

CS-Dialogue is publicly released under a CC BY-NC-SA 4.0 license for non-commercial academic research, enabling reproducible studies on Mandarin--English code-switching. The dataset page additionally provides a structured datasheet, the annotation guidelines, and the normalization and scoring scripts used in our evaluations.

\section{Potential Societal Impact}
This work may advance multilingual and code-switching ASR, enabling more inclusive speech technologies for diverse language users. Potential negative impacts include privacy risks and possible misuse in surveillance or speech processing systems. We mitigate these concerns by anonymizing all data and releasing the dataset under a non-commercial license with usage restrictions.

\section{Limitations}
While CS-Dialogue represents a significant contribution to the field, it has certain limitations.  First, the dataset focuses exclusively on Mandarin-English code-switching.  While this is a prevalent language pair, future work should expand to include other language combinations to enhance the generalizability of code-switching ASR models. Second, all participants are native Chinese speakers with strong English proficiency. The dataset does not include native English speakers who code-switch into Mandarin, which represents another important aspect of bilingual conversation. Third, although the dialogues are spontaneous, they were still recorded in a controlled environment, which may not fully reflect the acoustic diversity of real-world scenarios (e.g., noisy public spaces, varying microphone quality). Future work could explore data augmentation techniques to simulate a wider range of acoustic conditions. Fourth, exact per-speaker English-proficiency scores, overseas-exposure histories, and home-language or dialect backgrounds were not systematically retained during collection, which limits how finely the speaker population can be documented (see Appendix~\ref{appendix:speaker_demographics}).

\newpage
\section*{NeurIPS Paper Checklist}

\begin{enumerate}

\item {\bf Claims}
    \item[] Question: Do the main claims made in the abstract and introduction accurately reflect the paper's contributions and scope?
    \item[] Answer: \answerYes{} 
    \item[] Justification: The main claims in the abstract and introduction accurately reflect the paper’s contributions and scope. The paper introduces a new code-switching speech dataset, describes its construction and characteristics, and provides benchmark ASR results. These contributions are consistently supported by the corresponding sections in the paper without overclaiming. 
    \item[] Guidelines:
    \begin{itemize}
        \item The answer \answerNA{} means that the abstract and introduction do not include the claims made in the paper.
        \item The abstract and/or introduction should clearly state the claims made, including the contributions made in the paper and important assumptions and limitations. A \answerNo{} or \answerNA{} answer to this question will not be perceived well by the reviewers. 
        \item The claims made should match theoretical and experimental results, and reflect how much the results can be expected to generalize to other settings. 
        \item It is fine to include aspirational goals as motivation as long as it is clear that these goals are not attained by the paper. 
    \end{itemize}

\item {\bf Limitations}
    \item[] Question: Does the paper discuss the limitations of the work performed by the authors?
    \item[] Answer: \answerYes{} 
    \item[] Justification: The paper includes a dedicated "Limitations" section discussing dataset scope, potential bias, and challenges in code-switching ASR.
    \item[] Guidelines:
    \begin{itemize}
        \item The answer \answerNA{} means that the paper has no limitation while the answer \answerNo{} means that the paper has limitations, but those are not discussed in the paper. 
        \item The authors are encouraged to create a separate ``Limitations'' section in their paper.
        \item The paper should point out any strong assumptions and how robust the results are to violations of these assumptions (e.g., independence assumptions, noiseless settings, model well-specification, asymptotic approximations only holding locally). The authors should reflect on how these assumptions might be violated in practice and what the implications would be.
        \item The authors should reflect on the scope of the claims made, e.g., if the approach was only tested on a few datasets or with a few runs. In general, empirical results often depend on implicit assumptions, which should be articulated.
        \item The authors should reflect on the factors that influence the performance of the approach. For example, a facial recognition algorithm may perform poorly when image resolution is low or images are taken in low lighting. Or a speech-to-text system might not be used reliably to provide closed captions for online lectures because it fails to handle technical jargon.
        \item The authors should discuss the computational efficiency of the proposed algorithms and how they scale with dataset size.
        \item If applicable, the authors should discuss possible limitations of their approach to address problems of privacy and fairness.
        \item While the authors might fear that complete honesty about limitations might be used by reviewers as grounds for rejection, a worse outcome might be that reviewers discover limitations that aren't acknowledged in the paper. The authors should use their best judgment and recognize that individual actions in favor of transparency play an important role in developing norms that preserve the integrity of the community. Reviewers will be specifically instructed to not penalize honesty concerning limitations.
    \end{itemize}

\item {\bf Theory assumptions and proofs}
    \item[] Question: For each theoretical result, does the paper provide the full set of assumptions and a complete (and correct) proof?
    \item[] Answer: \answerNA{} 
    \item[] Justification: This paper proposes a new speech recognition dataset and does not include theoretical results.
    \item[] Guidelines:
    \begin{itemize}
        \item The answer \answerNA{} means that the paper does not include theoretical results. 
        \item All the theorems, formulas, and proofs in the paper should be numbered and cross-referenced.
        \item All assumptions should be clearly stated or referenced in the statement of any theorems.
        \item The proofs can either appear in the main paper or the supplemental material, but if they appear in the supplemental material, the authors are encouraged to provide a short proof sketch to provide intuition. 
        \item Inversely, any informal proof provided in the core of the paper should be complemented by formal proofs provided in appendix or supplemental material.
        \item Theorems and Lemmas that the proof relies upon should be properly referenced. 
    \end{itemize}

    \item {\bf Experimental result reproducibility}
    \item[] Question: Does the paper fully disclose all the information needed to reproduce the main experimental results of the paper to the extent that it affects the main claims and/or conclusions of the paper (regardless of whether the code and data are provided or not)?
    \item[] Answer: \answerYes{} 
    \item[] Justification: The proposed dataset is available on Huggingface now. The experimental details are also shown in the paper.
    \item[] Guidelines:
    \begin{itemize}
        \item The answer \answerNA{} means that the paper does not include experiments.
        \item If the paper includes experiments, a \answerNo{} answer to this question will not be perceived well by the reviewers: Making the paper reproducible is important, regardless of whether the code and data are provided or not.
        \item If the contribution is a dataset and\slash or model, the authors should describe the steps taken to make their results reproducible or verifiable. 
        \item Depending on the contribution, reproducibility can be accomplished in various ways. For example, if the contribution is a novel architecture, describing the architecture fully might suffice, or if the contribution is a specific model and empirical evaluation, it may be necessary to either make it possible for others to replicate the model with the same dataset, or provide access to the model. In general. releasing code and data is often one good way to accomplish this, but reproducibility can also be provided via detailed instructions for how to replicate the results, access to a hosted model (e.g., in the case of a large language model), releasing of a model checkpoint, or other means that are appropriate to the research performed.
        \item While NeurIPS does not require releasing code, the conference does require all submissions to provide some reasonable avenue for reproducibility, which may depend on the nature of the contribution. For example
        \begin{enumerate}
            \item If the contribuation is primarily a new algorithm, the paper should make it clear how to reproduce that algorithm.
            \item If the contribution is primarily a new model architecture, the paper should describe the architecture clearly and fully.
            \item If the contribution is a new model (e.g., a large language model), then there should either be a way to access this model for reproducing the results or a way to reproduce the model (e.g., with an open-source dataset or instructions for how to construct the dataset).
            \item We recognize that reproducibility may be tricky in some cases, in which case authors are welcome to describe the particular way they provide for reproducibility. In the case of closed-source models, it may be that access to the model is limited in some way (e.g., to registered users), but it should be possible for other researchers to have some path to reproducing or verifying the results.
        \end{enumerate}
    \end{itemize}

\item {\bf Open access to data and code}
    \item[] Question: Does the paper provide open access to the data and code, with sufficient instructions to faithfully reproduce the main experimental results, as described in supplemental material?
    \item[] Answer: \answerYes{} 
    \item[] Justification: The data is available at \url{https://huggingface.co/datasets/BAAI/CS-Dialogue}.
    \item[] Guidelines:
    \begin{itemize}
        \item The answer \answerNA{} means that paper does not include experiments requiring code.
        \item Please see the NeurIPS code and data submission guidelines (\url{https://neurips.cc/public/guides/CodeSubmissionPolicy}) for more details.
        \item While we encourage the release of code and data, we understand that this might not be possible, so \answerNo{} is an acceptable answer. Papers cannot be rejected simply for not including code, unless this is central to the contribution (e.g., for a new open-source benchmark).
        \item The instructions should contain the exact command and environment needed to run to reproduce the results. See the NeurIPS code and data submission guidelines (\url{https://neurips.cc/public/guides/CodeSubmissionPolicy}) for more details.
        \item The authors should provide instructions on data access and preparation, including how to access the raw data, preprocessed data, intermediate data, and generated data, etc.
        \item The authors should provide scripts to reproduce all experimental results for the new proposed method and baselines. If only a subset of experiments are reproducible, they should state which ones are omitted from the script and why.
        \item At submission time, to preserve anonymity, the authors should release anonymized versions (if applicable).
        \item Providing as much information as possible in supplemental material (appended to the paper) is recommended, but including URLs to data and code is permitted.
    \end{itemize}

\item {\bf Experimental setting/details}
    \item[] Question: Does the paper specify all the training and test details (e.g., data splits, hyperparameters, how they were chosen, type of optimizer) necessary to understand the results?
    \item[] Answer: \answerYes{} 
    \item[] Justification: See section 5 and appendix D.
    \item[] Guidelines:
    \begin{itemize}
        \item The answer \answerNA{} means that the paper does not include experiments.
        \item The experimental setting should be presented in the core of the paper to a level of detail that is necessary to appreciate the results and make sense of them.
        \item The full details can be provided either with the code, in appendix, or as supplemental material.
    \end{itemize}

\item {\bf Experiment statistical significance}
    \item[] Question: Does the paper report error bars suitably and correctly defined or other appropriate information about the statistical significance of the experiments?
    \item[] Answer: \answerNo{} 
    \item[] Justification: The paper reports deterministic metrics (CER, WER, MER) on fixed data splits, following standard practice in ASR benchmark and dataset work. We do not include error bars or statistical significance tests, as experiments are conducted without multiple random runs. While this limits variance estimation, our goal is to establish benchmark performance on the proposed dataset rather than compare closely competing methods.
    \item[] Guidelines:
    \begin{itemize}
        \item The answer \answerNA{} means that the paper does not include experiments.
        \item The authors should answer \answerYes{} if the results are accompanied by error bars, confidence intervals, or statistical significance tests, at least for the experiments that support the main claims of the paper.
        \item The factors of variability that the error bars are capturing should be clearly stated (for example, train/test split, initialization, random drawing of some parameter, or overall run with given experimental conditions).
        \item The method for calculating the error bars should be explained (closed form formula, call to a library function, bootstrap, etc.)
        \item The assumptions made should be given (e.g., Normally distributed errors).
        \item It should be clear whether the error bar is the standard deviation or the standard error of the mean.
        \item It is OK to report 1-sigma error bars, but one should state it. The authors should preferably report a 2-sigma error bar than state that they have a 96\% CI, if the hypothesis of Normality of errors is not verified.
        \item For asymmetric distributions, the authors should be careful not to show in tables or figures symmetric error bars that would yield results that are out of range (e.g., negative error rates).
        \item If error bars are reported in tables or plots, the authors should explain in the text how they were calculated and reference the corresponding figures or tables in the text.
    \end{itemize}

\item {\bf Experiments compute resources}
    \item[] Question: For each experiment, does the paper provide sufficient information on the computer resources (type of compute workers, memory, time of execution) needed to reproduce the experiments?
    \item[] Answer: \answerYes{} 
    \item[] Justification: Compute resources, including GPU type and training setup, are described in Appendix D.
    \item[] Guidelines:
    \begin{itemize}
        \item The answer \answerNA{} means that the paper does not include experiments.
        \item The paper should indicate the type of compute workers CPU or GPU, internal cluster, or cloud provider, including relevant memory and storage.
        \item The paper should provide the amount of compute required for each of the individual experimental runs as well as estimate the total compute. 
        \item The paper should disclose whether the full research project required more compute than the experiments reported in the paper (e.g., preliminary or failed experiments that didn't make it into the paper). 
    \end{itemize}
    
\item {\bf Code of ethics}
    \item[] Question: Does the research conducted in the paper conform, in every respect, with the NeurIPS Code of Ethics \url{https://neurips.cc/public/EthicsGuidelines}?
    \item[] Answer: \answerYes{} 
    \item[] Justification: The data collection and usage comply with the NeurIPS Code of Ethics. All participants provided informed consent, and the data was anonymized to remove any personally identifiable information. The dataset is released for non-commercial academic use only, and ethical considerations are discussed in the paper. 
    \item[] Guidelines:
    \begin{itemize}
        \item The answer \answerNA{} means that the authors have not reviewed the NeurIPS Code of Ethics.
        \item If the authors answer \answerNo, they should explain the special circumstances that require a deviation from the Code of Ethics.
        \item The authors should make sure to preserve anonymity (e.g., if there is a special consideration due to laws or regulations in their jurisdiction).
    \end{itemize}

\item {\bf Broader impacts}
    \item[] Question: Does the paper discuss both potential positive societal impacts and negative societal impacts of the work performed?
    \item[] Answer: \answerYes{} 
    \item[] Justification: Potential societal impacts are discussed in the appendix, including applications in multilingual ASR and risks related to privacy and misuse.
    \item[] Guidelines:
    \begin{itemize}
        \item The answer \answerNA{} means that there is no societal impact of the work performed.
        \item If the authors answer \answerNA{} or \answerNo, they should explain why their work has no societal impact or why the paper does not address societal impact.
        \item Examples of negative societal impacts include potential malicious or unintended uses (e.g., disinformation, generating fake profiles, surveillance), fairness considerations (e.g., deployment of technologies that could make decisions that unfairly impact specific groups), privacy considerations, and security considerations.
        \item The conference expects that many papers will be foundational research and not tied to particular applications, let alone deployments. However, if there is a direct path to any negative applications, the authors should point it out. For example, it is legitimate to point out that an improvement in the quality of generative models could be used to generate Deepfakes for disinformation. On the other hand, it is not needed to point out that a generic algorithm for optimizing neural networks could enable people to train models that generate Deepfakes faster.
        \item The authors should consider possible harms that could arise when the technology is being used as intended and functioning correctly, harms that could arise when the technology is being used as intended but gives incorrect results, and harms following from (intentional or unintentional) misuse of the technology.
        \item If there are negative societal impacts, the authors could also discuss possible mitigation strategies (e.g., gated release of models, providing defenses in addition to attacks, mechanisms for monitoring misuse, mechanisms to monitor how a system learns from feedback over time, improving the efficiency and accessibility of ML).
    \end{itemize}
    
\item {\bf Safeguards}
    \item[] Question: Does the paper describe safeguards that have been put in place for responsible release of data or models that have a high risk for misuse (e.g., pre-trained language models, image generators, or scraped datasets)?
    \item[] Answer: \answerYes{} 
    \item[] Justification: The dataset is released with anonymization and usage restrictions (non-commercial license), and all personally identifiable information has been removed.
    \item[] Guidelines:
    \begin{itemize}
        \item The answer \answerNA{} means that the paper poses no such risks.
        \item Released models that have a high risk for misuse or dual-use should be released with necessary safeguards to allow for controlled use of the model, for example by requiring that users adhere to usage guidelines or restrictions to access the model or implementing safety filters. 
        \item Datasets that have been scraped from the Internet could pose safety risks. The authors should describe how they avoided releasing unsafe images.
        \item We recognize that providing effective safeguards is challenging, and many papers do not require this, but we encourage authors to take this into account and make a best faith effort.
    \end{itemize}

\item {\bf Licenses for existing assets}
    \item[] Question: Are the creators or original owners of assets (e.g., code, data, models), used in the paper, properly credited and are the license and terms of use explicitly mentioned and properly respected?
    \item[] Answer: \answerYes{} 
    \item[] Justification: The dataset is released under a CC BY-NC-SA license, and all used assets are properly credited with their respective licenses.
    \item[] Guidelines:
    \begin{itemize}
        \item The answer \answerNA{} means that the paper does not use existing assets.
        \item The authors should cite the original paper that produced the code package or dataset.
        \item The authors should state which version of the asset is used and, if possible, include a URL.
        \item The name of the license (e.g., CC-BY 4.0) should be included for each asset.
        \item For scraped data from a particular source (e.g., website), the copyright and terms of service of that source should be provided.
        \item If assets are released, the license, copyright information, and terms of use in the package should be provided. For popular datasets, \url{paperswithcode.com/datasets} has curated licenses for some datasets. Their licensing guide can help determine the license of a dataset.
        \item For existing datasets that are re-packaged, both the original license and the license of the derived asset (if it has changed) should be provided.
        \item If this information is not available online, the authors are encouraged to reach out to the asset's creators.
    \end{itemize}

\item {\bf New assets}
    \item[] Question: Are new assets introduced in the paper well documented and is the documentation provided alongside the assets?
    \item[] Answer: \answerYes{} 
    \item[] Justification: The dataset is well documented, including data collection, annotation, and usage details.
    \item[] Guidelines:
    \begin{itemize}
        \item The answer \answerNA{} means that the paper does not release new assets.
        \item Researchers should communicate the details of the dataset\slash code\slash model as part of their submissions via structured templates. This includes details about training, license, limitations, etc. 
        \item The paper should discuss whether and how consent was obtained from people whose asset is used.
        \item At submission time, remember to anonymize your assets (if applicable). You can either create an anonymized URL or include an anonymized zip file.
    \end{itemize}

\item {\bf Crowdsourcing and research with human subjects}
    \item[] Question: For crowdsourcing experiments and research with human subjects, does the paper include the full text of instructions given to participants and screenshots, if applicable, as well as details about compensation (if any)? 
    \item[] Answer: \answerYes{}
    \item[] Justification: 
    The paper describes the data collection process, including participant instructions, recording setup, and compensation details.
    \item[] Guidelines:
    \begin{itemize}
        \item The answer \answerNA{} means that the paper does not involve crowdsourcing nor research with human subjects.
        \item Including this information in the supplemental material is fine, but if the main contribution of the paper involves human subjects, then as much detail as possible should be included in the main paper. 
        \item According to the NeurIPS Code of Ethics, workers involved in data collection, curation, or other labor should be paid at least the minimum wage in the country of the data collector. 
    \end{itemize}

\item {\bf Institutional review board (IRB) approvals or equivalent for research with human subjects}
    \item[] Question: Does the paper describe potential risks incurred by study participants, whether such risks were disclosed to the subjects, and whether Institutional Review Board (IRB) approvals (or an equivalent approval/review based on the requirements of your country or institution) were obtained?
    \item[] Answer: \answerYes{} 
    \item[] Justification: All participants provided informed consent, and the data collection followed institutional and ethical guidelines. Personal information was removed to ensure privacy.
    \item[] Guidelines:
    \begin{itemize}
        \item The answer \answerNA{} means that the paper does not involve crowdsourcing nor research with human subjects.
        \item Depending on the country in which research is conducted, IRB approval (or equivalent) may be required for any human subjects research. If you obtained IRB approval, you should clearly state this in the paper. 
        \item We recognize that the procedures for this may vary significantly between institutions and locations, and we expect authors to adhere to the NeurIPS Code of Ethics and the guidelines for their institution. 
        \item For initial submissions, do not include any information that would break anonymity (if applicable), such as the institution conducting the review.
    \end{itemize}

\item {\bf Declaration of LLM usage}
    \item[] Question: Does the paper describe the usage of LLMs if it is an important, original, or non-standard component of the core methods in this research? Note that if the LLM is used only for writing, editing, or formatting purposes and does \emph{not} impact the core methodology, scientific rigor, or originality of the research, declaration is not required.
    \item[] Answer: \answerNA 
    \item[] Justification: LLMs are not used as part of the core methodology of this work.
    \item[] Guidelines:
    \begin{itemize}
        \item The answer \answerNA{} means that the core method development in this research does not involve LLMs as any important, original, or non-standard components.
        \item Please refer to our LLM policy in the NeurIPS handbook for what should or should not be described.
    \end{itemize}

\end{enumerate}

\end{document}